\documentclass[journal]{IEEEtran}
\ifCLASSINFOpdf
\else
\fi
\usepackage{url}
\usepackage{bm}
\usepackage[utf8]{inputenc}
\usepackage[numbers,sort&compress]{natbib}
\usepackage{booktabs}
\usepackage{scalerel}
\usepackage{tikz}
\usetikzlibrary{svg.path}
\usepackage{graphicx}
\usepackage{tikz,xcolor,hyperref}
\usepackage{subfigure}
\usepackage{multirow}
\usepackage{color}
\usepackage{float}
\usepackage{footnote}
\usepackage{stfloats}
\usepackage{array}
\usepackage{booktabs}
\usepackage[noend]{algpseudocode}
\usepackage{amsmath} 
\usepackage[ruled,vlined]{algorithm2e}
 \usepackage{amsfonts} 
 \usepackage{url}
 \usepackage{balance}
 \usepackage[misc]{ifsym} 
\usepackage{booktabs}
\usepackage{multirow}
\usepackage{graphicx}
\usepackage[normalem]{ulem}
\useunder{\uline}{\ul}{}
\usepackage{amssymb}

\SetAlgoSkip{15pt} 
\SetAlgoInsideSkip{15pt} 

\usepackage{xcolor}
\newcommand{\hys}[1][\textcolor{black}]{#1}
\usepackage{xcolor}
\newcommand{\hyss}[1][\textcolor{black}]{#1}
\newcommand{\tgrs}[1][\textcolor{black}]{#1}

\newcommand{\eg}{\textit{e.g.}}

\UseRawInputEncoding

\newcommand{\RNum}[1]{\uppercase\expandafter{\romannumeral #1\relax}}

\definecolor{lime}{HTML}{A6CE39}
\DeclareRobustCommand{\orcidicon}{%
	\begin{tikzpicture}
	\draw[lime, fill=lime] (0,0) 
	circle [radius=0.16] 
	node[white] {{\fontfamily{qag}\selectfont \tiny ID}};
	\draw[white, fill=white] (-0.0625,0.095) 
	circle [radius=0.007];
	\end{tikzpicture}
	\hspace{-2mm}
}

\foreach \x in {A, ..., Z}{%
	\expandafter\xdef\csname orcid\x\endcsname{\noexpand\href{https://orcid.org/\csname orcidauthor\x\endcsname}{\noexpand\orcidicon}}
}

\begin{document}
%
\title{IRSRMamba: Infrared Image Super-Resolution via Mamba-based Wavelet Transform Feature Modulation Model}
%
%
%

\author{Yongsong~Huang\orcidA{}~\IEEEmembership{Member,~IEEE}, Tomo Miyazaki~\IEEEmembership{Member,~IEEE}, Xiaofeng Liu~\IEEEmembership{Member,~IEEE}, Shinichiro Omachi~\IEEEmembership{Senior Member,~IEEE} 

\thanks{This work was supported by JSPS KAKENHI Grant Number JP23KJ0118. (\textit{Corresponding author:} Yongsong Huang$^{\textrm{\Letter}}$.) Yongsong Huang is with the Graduate School of Engineering Tohoku University (and Yale University), Sendai 9808579, Japan. (e-mail: hys@dc.tohoku.ac.jp)}

\thanks{Tomo Miyazaki and Shinichiro Omachi with Graduate School of Engineering, Tohoku University, Sendai 9808579, Japan. Xiaofeng Liu is with Yale University, New Haven 06519, USA.}
}

%
%

\markboth{Journal of \LaTeX\ Class Files,~Vol.~13, No.~9, September~2014}%
{Shell \MakeLowercase{\textit{et al.}}: Bare Demo of IEEEtran.cls for Journals}
%



\maketitle

\begin{abstract}
Infrared image super-resolution demands long-range dependency modeling and multi-scale feature extraction to address challenges such as homogeneous backgrounds, weak edges, and sparse textures. While Mamba-based state-space models (SSMs) excel in global dependency modeling with linear complexity, their block-wise processing disrupts spatial consistency, limiting their effectiveness for IR image reconstruction. We propose IRSRMamba, a novel framework integrating wavelet transform feature modulation for multi-scale adaptation and an \tgrs{SSMs-based semantic consistency loss} to restore fragmented contextual information. This design enhances global-local feature fusion, structural coherence, and fine-detail preservation while mitigating block-induced artifacts. Experiments on benchmark datasets demonstrate that IRSRMamba outperforms state-of-the-art methods in PSNR, SSIM, and perceptual quality. This work establishes Mamba-based architectures as a promising direction for high-fidelity IR image enhancement. Code are available at \url{https://github.com/yongsongH/IRSRMamba}.
\end{abstract}


%
\IEEEpeerreviewmaketitle

\section{Introduction} \IEEEPARstart{I}{nfrared} imaging is widely applied in security surveillance, remote sensing, and interplanetary exploration, where visible-light imaging is often ineffective due to environmental constraints\cite{harvey2012first}. However, infrared (IR) images often suffer from low spatial resolution, noise contamination, and contrast degradation, primarily due to sensor limitations and atmospheric interference. To address these issues, single-image super-resolution (SISR) has emerged as a promising computational approach, reconstructing high-resolution (HR) images from low-resolution (LR) counterparts using deep learning models\cite{huang2021infrared, jiang2021difference, zhang2024joint}.

Despite recent advancements in SISR, infrared image super-resolution (IRSR) remains particularly challenging due to the unique statistical properties of IR images. Unlike natural images, which typically exhibit well-defined structures and rich textures, IR images often feature homogeneous backgrounds, weak edge contrast, and sparse high-frequency details, making it difficult for conventional models to reconstruct fine details while maintaining global coherence\cite{chen2023efficient, zhou2022structure}. While CNN- and Transformer-based IRSR methods have demonstrated promising results, they struggle with efficient long-range dependency modeling and fail to preserve intricate textures, particularly in infrared scenes with non-uniform feature distributions\cite{huang2021infrared, huang2023infrared, ma2023msma, zhao2024modality}.

Recently, state-space models (SSMs) have emerged as an efficient alternative to traditional deep learning architectures, with Mamba demonstrating strong performance in image segmentation, restoration, and large-scale vision tasks\cite{gu2023mamba, guo2024mambair, zhu2024vision}. Unlike CNNs and Transformers, which rely on localized convolutions or computationally expensive self-attention mechanisms, Mamba efficiently captures long-range dependencies with linear complexity, making it well-suited for large-scale image processing\cite{zhang2024survey, zhu2024vision, wang2024state}. However, applying Mamba directly to IRSR introduces new challenges. Since Mamba processes images in a block-wise manner, it disrupts spatial consistency, as infrared images exhibit highly non-uniform structural distributions\cite{Yu2024Rethinking}. This partitioning leads to fragmented global context, limiting the model’s ability to recover fine textures across different regions. Additionally, in IR images, where salient structures are sparse, adjacent blocks may encode correlated features yet lack direct information exchange, further complicating detail reconstruction.

\begin{figure}[t]
\centerline{\includegraphics[width=\columnwidth]{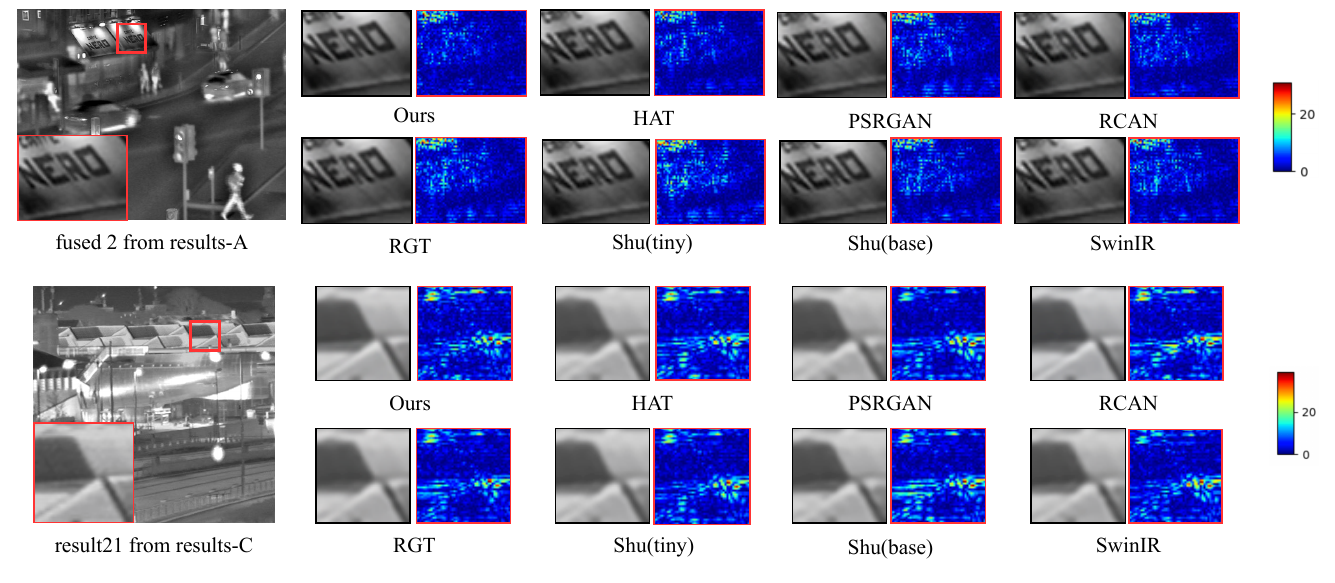}}
\caption{\tgrs{Super-resolution results on benchmark infrared datasets result-A and result-C at scaling factors of $\times 2$ and $\times 4$. The left panels display the SR images, while the right panels present the corresponding residual visualizations, with color bars indicating the degree of residuals-warmer colors denote larger discrepancies. }\textbf{Top} The SR results of fused2 from results-A with scale factor of $\times 2$.  \textbf{Below} The SR results of result21 from results-C with scale factor of $\times 4$.}
\label{fig.5}
\end{figure}

To address these challenges, we introduce IRSRMamba, the first Mamba-based IRSR framework specifically designed to mitigate global information fragmentation caused by block-wise processing. Our approach integrates wavelet transform feature modulation, which improves multi-scale receptive field adaptation by compensating for spatial coherence loss while refining high-frequency details. Unlike prior methods that apply wavelet transforms primarily for denoising or auxiliary enhancement\cite{xue2024low, jiang2023low, liu2023mwln}, our method seamlessly integrates wavelet-based modulation into the Mamba feature learning pipeline, ensuring multi-scale feature interaction and global-local information fusion. By encoding frequency-domain representations, IRSRMamba effectively reconstructs both fine textures and large-scale structures, mitigating the limitations introduced by block-based representation.

Beyond wavelet-based enhancements, we introduce an \tgrs{SSMs-based semantic consistency loss}, which enforces cross-block feature alignment to maintain structural coherence. This loss, detailed in Algorithm .\ref{al.1}, applies directional sequence modeling to preserve semantic integrity, ensuring that related features remain consistent across different regions. By jointly leveraging wavelet-driven multi-scale feature extraction and state-space semantic constraints, IRSRMamba not only surpasses existing IRSR methods in perceptual fidelity and structural accuracy but also establishes a new paradigm for infrared image reconstruction within state-space modeling frameworks.

To further assess the impact of our method, we employ residual visualization for error analysis. As demonstrated in Figure \ref{fig.5}, IRSRMamba exhibits significantly reduced reconstruction errors compared to existing approaches, highlighting its superior ability to restore fine-grained details and maintain structural consistency. The residual maps illustrate how our model effectively recovers high-frequency textures while maintaining global coherence, validating the effectiveness of integrating state-space modeling with frequency-domain enhancement in IRSR.

In summary, the main contributions of our work are as follows:

\begin{itemize}
    \item We introduce Mamba as a backbone for IRSR, leveraging its state-space modeling capabilities to efficiently capture long-range dependencies, significantly improving the reconstruction of sparse infrared textures.
    \item We propose a wavelet-driven feature modulation mechanism that enhances multi-scale receptive field adaptation, effectively capturing both spatial and frequency-domain representations, thereby mitigating fragmentation caused by block-wise processing.
    \item To address the global information fragmentation resulting from block-wise modeling, we introduce an \tgrs{SSMs-based semantic consistency loss}, enforcing cross-block alignment to enhance structural coherence and fine-detail restoration.
    \item Extensive experiments demonstrate that IRSRMamba outperforms state-of-the-art IRSR models, achieving superior pixel-wise fidelity and perceptual quality, with ablation studies validating the effectiveness of each component.
\end{itemize}

\section{Related Work}

\subsection{Infrared Image Super-Resolution \& Block-wise Processing}

Infrared image super-resolution remains a significant challenge due to the low spatial resolution, high noise levels, and reduced contrast inherent in infrared imaging. These limitations stem from sensor constraints, environmental interference, and the fundamental physics of infrared light capture. Traditional model-based approaches have sought to mitigate these effects by incorporating hand-crafted priors and physical constraints, but their rigidity prevents them from effectively adapting to the complex degradations and diverse imaging conditions encountered in real-world scenarios.

With the rise of deep learning, substantial progress has been made in IRSR, with CNN-, GAN-, and Transformer-based models demonstrating remarkable success. CNNs leverage hierarchical feature extraction to enhance spatial details, while GANs introduce adversarial training strategies to improve perceptual realism\cite{li2019feedback,huang2021infrared}. Transformer-based architectures extend these capabilities by capturing long-range dependencies and modeling global contextual information, which is essential for high-fidelity restoration\cite{shi2024swinibsr,qin2024lkformer,wu2023swinipisr}. Additionally, hybrid approaches, such as multimodal data fusion (\eg, visible-infrared joint reconstruction) and physics-informed priors, have further improved robustness against domain-specific degradations\cite{zhang2024dual,wu2023deep,suarez2024enhancement}.

Despite these advancements, block-wise processing remains a major limitation in IRSR. Modern super-resolution methods, particularly Transformers and SSMs, process patch-based images to reduce computational complexity. However, this introduces spatial fragmentation, leading to inconsistencies in global structure and texture coherence. Infrared images, characterized by non-uniform structural distributions and sparse textures, are particularly vulnerable to this issue, as adjacent patches often contain strongly correlated features yet lack explicit cross-block information exchange. This challenge can be formulated mathematically by defining the image as a collection of feature blocks:

\begin{equation}
\mathbf{X}_b=\left\{\mathbf{x}_i \mid i \in \mathcal{I}_b\right\}, \quad \forall b \in\{1, \ldots, B\}
\end{equation} where $\mathbf{X}_b$ represents the feature set of the $b$-th block, and $\mathcal{I}_b$ denotes the pixel indices contained within that block. Many block-based methods assume that the joint probability distribution of the full image features can be factorized across blocks:

\begin{equation}
P(\mathbf{X} \mid \Theta)=\prod_{b=1}^B P\left(\mathbf{X}_b \mid \Theta_b\right)
\end{equation} where $P(\mathbf{X} \mid \Theta)$ is the joint probability distribution of the full image features given model parameters $\Theta$, but its factorization across independent blocks disrupts the assumption of global spatial coherence.

The block-wise fragmentation effect can be further quantified through the expected reconstruction error across blocks:

\begin{equation}\resizebox{0.9\hsize}{!}{$E\left[\left|I_{\mathrm{HR}}-I_{\mathrm{SR}}\right|^2\right] \approx \sum_{b=1}^{B}  E\left[\left|\mathbf{X}_b-\mathbf{X}_b^{\mathrm{SR}}\right|^2\right]+\sum_{i \neq j} E\left[\mathbf{X}_i \cdot \mathbf{X}_j\right] $} \end{equation} where $I_{\mathrm{HR}}$ and $I_{\mathrm{SR}}$ represent the high-resolution ground truth and SR infrared images, respectively. The second summation term captures cross-block feature dependencies, which conventional block-wise models fail to incorporate. $E[\cdot]$ denotes the expectation operator over the data distribution. For infrared images, this residual is particularly large due to their low-frequency dominant nature, leading to structural artifacts and texture misalignment in reconstructed images. This phenomenon exacerbates domain gaps, making it difficult for supervised deep-learning models to generalize across different imaging conditions\cite{Yu2024Rethinking}.



To address the fragmentation induced by block-wise processing, we propose IRSRMamba, an infrared image super-resolution framework that integrates wavelet transform feature modulation (WTFM) and an SSMs-based semantic consistency loss. Unlike conventional Transformer-based IRSR methods that rely on static tokenization schemes, IRSRMamba employs Wavelet Transform Feature Modulation to dynamically adjust receptive fields, ensuring that critical high-frequency details are preserved while maintaining large-scale contextual integrity. By encoding frequency-domain representations directly into a Mamba-based state-space model, IRSRMamba improves feature fusion across different resolutions, mitigating aliasing and discontinuities introduced by block-wise processing. Additionally, the SSMs-based semantic consistency loss explicitly enforces feature alignment across fragmented blocks, ensuring spatial consistency throughout the reconstructed image. This loss function is designed to minimize inter-block feature variance, reducing perceptual distortions caused by block-based partitioning.

\subsection{Mamba-Based Models in Image Processing} Recent developments in SSMs have led to the introduction of Mamba, an architecture designed for long-range dependency modeling with linear computational complexity\cite{gu2023mamba}. Traditional deep learning architectures exhibit inherent limitations—CNNs struggle with restricted receptive fields, while Transformers incur quadratic complexity in self-attention operations. Mamba mitigates these inefficiencies by leveraging structured state-space representations, which effectively capture both global and local contextual dependencies\cite{zhang2024survey, wang2024state}.

In image processing, Mamba has been successfully applied to image restoration, super-resolution, and classification. MambaIR\cite{guo2024mambair} establishes a robust baseline for image restoration by integrating SSM-based sequence modeling with domain-specific priors, improving feature consistency and structural preservation. Meanwhile, Vision Mamba\cite{zhu2024vision} extends Mamba’s applicability to visual representation learning, demonstrating competitive performance in bidirectional sequence modeling.

For infrared imaging, Mamba-based architectures have been explored in small-target detection and hyperspectral image analysis. MiM-ISTD\cite{chen2024mim} introduces a hierarchical Mamba-in-Mamba structure for infrared small-target detection, efficiently capturing fine-grained local details and global spatial structures. Similarly, MambaHSI\cite{li2024mambahsi} adapts Mamba for hyperspectral image classification, integrating spatial-spectral feature modeling to overcome the computational inefficiencies associated with Transformer-based approaches. Additionally, IGroupSS-Mamba\cite{he2024igroupss} employs a hierarchical grouping strategy to enhance spatial-spectral feature extraction, making it particularly effective for high-dimensional remote sensing data.


\subsection{Wavelet Transform} Wavelet transform has been widely adopted in image processing for its ability to decompose images into multi-resolution frequency components while preserving spatial locality. Traditional wavelet-based methods focus on multi-scale representations to enhance feature extraction, making them particularly effective for tasks requiring fine-grained structural preservation and global contextual awareness. More recently, wavelet-based convolutions have been introduced to extend receptive fields in CNNs while maintaining computational efficiency, enabling multi-frequency response without excessive parameter growth\cite{finder2025wavelet}.

In image restoration, wavelet transform has proven highly effective in contrast enhancement, noise suppression, and high-frequency detail preservation. By decomposing images into different frequency bands, wavelet-based models apply targeted enhancements to high-frequency details, while maintaining global structural integrity\cite{tan2024wavelet}. This approach has been particularly beneficial in infrared imaging, where traditional convolution-based super-resolution models struggle to recover fine textures and intricate patterns.

Recent studies have further integrated wavelet decomposition with deep learning architectures, particularly Mamba-based models, enhancing feature extraction efficiency while overcoming the limitations of CNNs and Transformers. The combination of wavelet-based processing with state-space modeling enables effective long-range dependency modeling while maintaining computational efficiency\cite{peng2025wmamba}. These advancements have opened new possibilities for hybrid architectures, incorporating wavelet-domain enhancements and domain-specific priors to optimize image reconstruction in medical imaging, remote sensing, and security surveillance.


\section{Methodology}

\subsection{Preliminaries}
\textbf{IR Image Super-Resolution:} In the field of IRSR, current methodologies utilize paired training datasets denoted by $\left\{x_{ir}, y_{ir}\right\}$. Here, $x_{ir} \in \mathbb{R}^{\frac{H}{\mu} \times \frac{W}{\mu} \times C}$ represents the low-resolution IR images, and $y_{ir} \in \mathbb{R}^{H \times W \times C}$ represents the corresponding high-resolution IR images, with $\mu$ indicating the spatial upscaling factor that affects both the height ($H$) and width ($W$) dimensions. Here, $C$ denotes the channel count of RGB images, typically three. \textbf{State Space Models:} Structured state-space models (SSMs) have recently gained traction in sequence modeling due to their ability to efficiently capture long-range dependencies. Inspired by continuous linear time-invariant (LTI) systems, SSMs map a one-dimensional sequence $x(t)$ to an output $y(t)$ via an implicit latent state $h(t)$, governed by the following differential equations:

\begin{equation}
\begin{aligned}
h(t) & =\mathbf{A} h(t-1)+\mathbf{B} x(t) \\
y(t) & =\mathbf{C} h(t)+\mathbf{D} x(t)
\end{aligned}
\end{equation}
where $\mathbf{A} \in \mathbb{R}^{N \times N}, \mathbf{B} \in \mathbb{R}^{N \times 1}, \mathbf{C} \in \mathbb{R}^{1 \times N}$, and $\mathbf{D} \in \mathbb{R}$ define the system parameters. To integrate SSMs into deep learning frameworks, these continuous parameters are discretized using the zero-order hold $(\mathrm{ZOH})$ method with a timescale parameter $\Delta$, yielding:

\begin{equation}
\begin{aligned}
& \overline{\mathbf{A}}=\exp (\Delta \mathbf{A}) \\
& \overline{\mathbf{B}}=(\Delta \mathbf{A})^{-1}(\exp (\mathbf{A})-\mathbf{I}) \cdot \Delta \mathbf{B}
\end{aligned}
\end{equation}

The discretized SSM formulation, suitable for recurrent computations, is given by:

\begin{equation}
\begin{aligned}
h_t & =\overline{\mathbf{A}} h_{t-1}+\overline{\mathbf{B}} x_t \\
y_t & =\mathbf{C} h_t+\mathbf{D} x_t
\end{aligned}
\end{equation} where $h_t$ captures sequential dependencies, making it well-suited for long-range modeling in image processing tasks. Furthermore, these equations can be transformed into a CNN compatible framework, where the convolution operation is defined as:

\begin{equation}
\resizebox{0.6\columnwidth}{!}{$
\begin{aligned}
\overline{\mathbf{K}} & \triangleq\left(\mathbf{C} \overline{\mathbf{B}}, \mathbf{C} \overline{\mathbf{A B}}, \cdots, \mathbf{C A}^{L-1} \overline{\mathbf{B}}\right), \\
\mathbf{y} & =\mathbf{x} \circledast \overline{\mathbf{K}}
\end{aligned}
$}
\end{equation}
where $L$ denotes the input sequence length, and $\overline{\mathbf{K}} \in \mathbb{R}^L$ serves as a structured convolution kernel. The convolution operation $\circledast$ enables efficient feature extraction over spatially structured data.

\begin{figure*}[htbp]
\centerline{\includegraphics[width=0.85\textwidth]{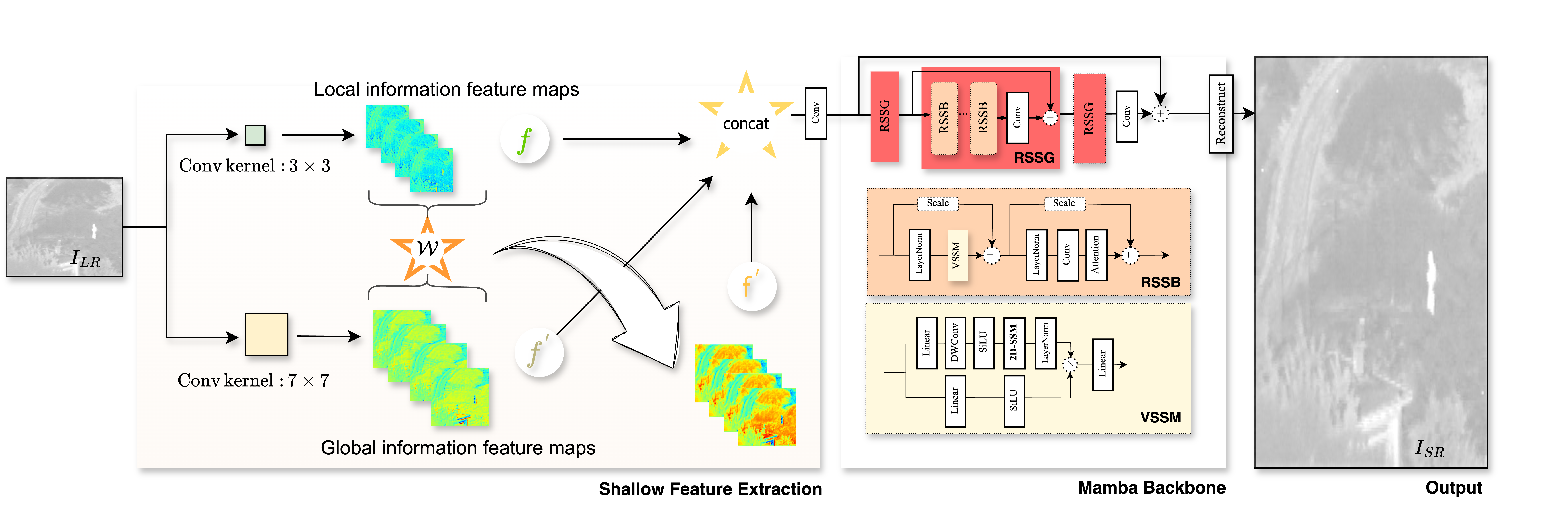}}
\caption{The architecture of IRSRMamba with the input image data $I_{LR}\in x_{ir}$ and $concat$ denotes the concat operation. IRSRMamba model includes shallow feature extraction, Mamba backbone network, and reconstruction tail.  For shallow feature extraction: there are three different feature maps shown here: the small receptive field feature map $f$ from the $3 \times 3$ convolutional kernel, the large receptive field $7 \times 7$ feature map $f^{\prime}$ and the wavelet transform modulated feature map defined as $\mathsf{f}^{'}$. The Wavelet Transform Feature Modulation Block $\mathcal{W}$ \hyss{is shown in} Fig.\ref{fig.2}. And the Mamba backbone network is based on \hys{Residual State-Space Groups (RSSG), which include the Residual State-Space Block (RSSB),} the Vision State-Space Module (VSSM), and the 2D Selective Scan Module (2D-SSM, see Fig.\ref{fig.3}). Best viewed in color. }
\label{fig.1}
\vspace{-10pt}
\end{figure*}

\begin{figure}[]
\centerline{\includegraphics[width=\columnwidth]{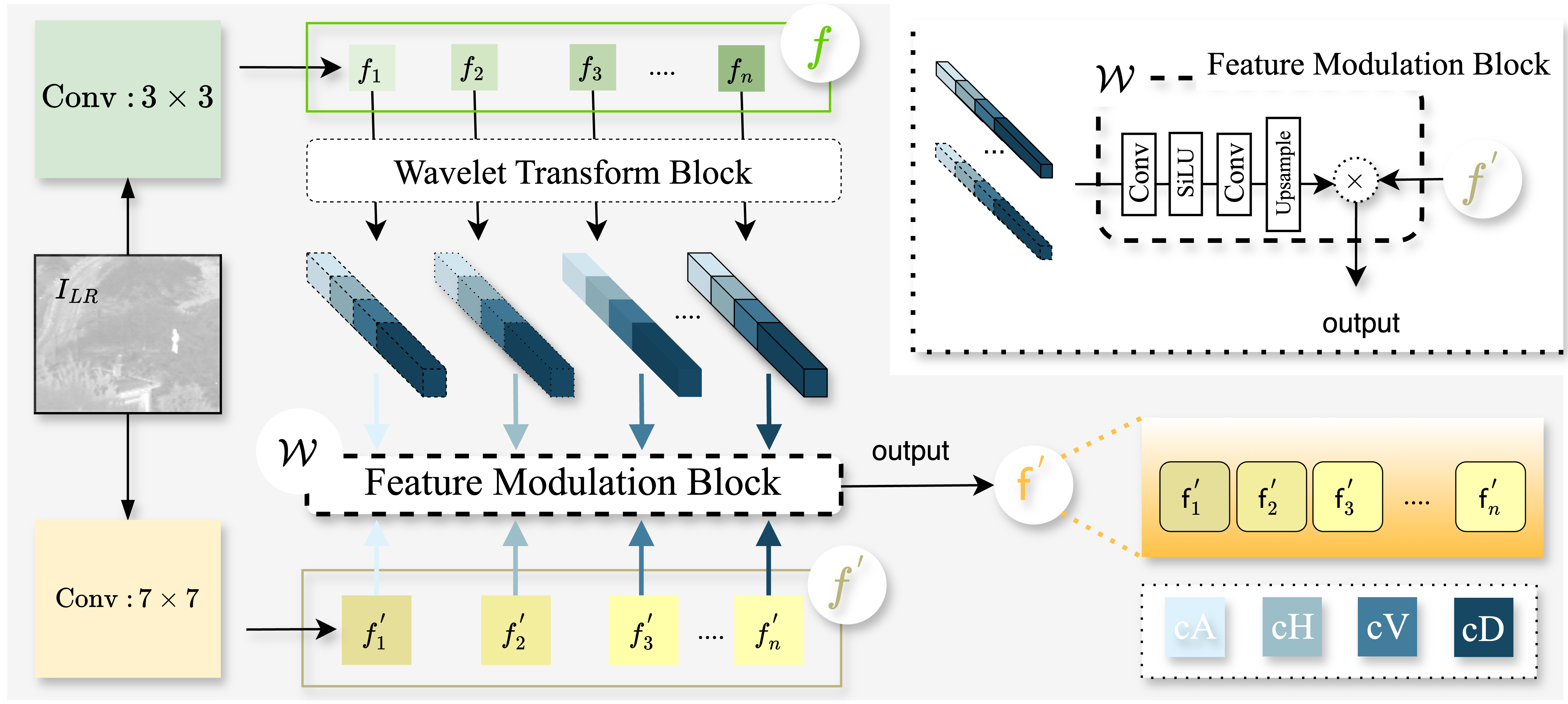}}
\caption{The architecture of Wavelet Transform Feature Modulation Block. $n$ denotes the feature maps of different channels, and feature modulation happens between the corresponding feature maps.}
\label{fig.2}
\vspace{-10pt}
\end{figure}

\begin{figure}[]
\centerline{\includegraphics[width=0.9\columnwidth]{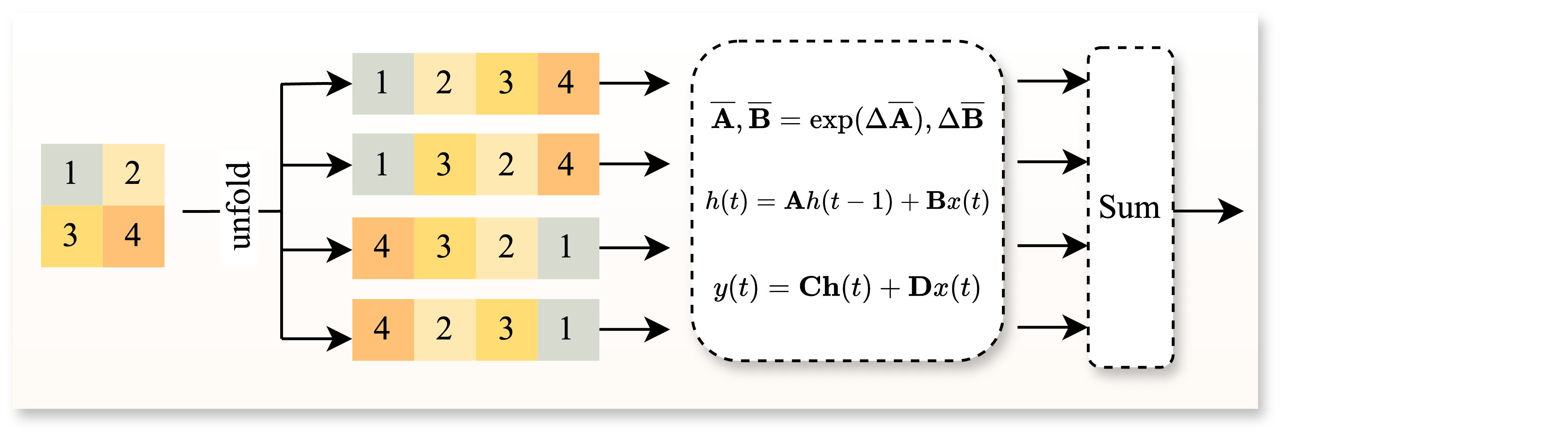}}
\caption{The architecture of 2D-Selective Scan Module (2D-SSM).}
\label{fig.3}
\vspace{-10pt}
\end{figure}
\vspace{-10pt}

\subsection{IRSRMamba Architecture}

IRSRMamba pioneers the application of Mamba-based state-space modeling in IRSR, enabling efficient contextual dependency modeling for enhanced texture restoration. As shown in Fig. \ref{fig.1}, IRSRMamba consists of three key components: \textbf{Shallow Feature Extraction.} The input LR image $I_{L R}$ is processed through a $3 \times 3$ convolution, yielding initial feature representations. These are further expanded using both a $7 \times 7$ convolutional kernel for broader context awareness and a Wavelet Transform Feature Modulation Block $\mathcal{W}$ (Fig. \ref{fig.2}), which enhances feature representations with multi-scale frequency-domain details. \textbf{Mamba Backbone Network.} Deep feature extraction is performed through Residual State Space Groups (RSSG), each containing multiple Residual State-Space Blocks (RSSB). RSSB replaces traditional Transformer attention layers with Vision State-Space Modules (VSSM) and LayerNorm (LN), effectively capturing spatial long-range dependencies. Additionally, RSSB incorporates a learnable scale factor $s$, dynamically adjusting skip connections:

\begin{equation}
Z^i=\operatorname{VSSM}\left(\operatorname{LN}\left(F_D^i\right)\right)+s \cdot F_D^i,
\end{equation} where $F_D^i$ represents the deep feature input at layer $i$, and $Z^i$ is the refined feature output.

\textbf{Reconstruction Tail.} The processed features are passed through additional convolutional layers to reconstruct the HR output $I_{HR}$, ensuring faithful texture recovery and structural integrity.

\subsection{Wavelet Transform Feature Modulation Block}
 
The Wavelet Transform Feature Modulation Block $\mathcal{W}$ enhances shallow feature extraction by leveraging multi-scale frequency representations. This block initiates with an input \( x_{ir} \). The modulation process begins with the application of two distinct convolutional modules, $\operatorname{Conv}_{3 \times 3}$ and $\operatorname{Conv}_{7 \times 7}$, which extract features at different scales:
\begin{equation}
\begin{aligned}
& f=\operatorname{Conv}_{3 \times 3}\left(x_{i r}\right) \\
& f^{\prime}=\operatorname{Conv}_{7 \times 7}\left(x_{i r}\right)
\end{aligned}
\end{equation}
Here, \( f \) captures finer details using a \( 3 \times 3 \) convolution kernel, while \( f' \) secures broader contextual information via a \( 7 \times 7 \) kernel. Subsequently, the small-scale feature map \( f \) is processed through the \hys{wavelet transform block (WTB),} $c A, c H, c V, c D=\operatorname{WTB}(f)$ resulting in four components that detail various directional textures:
\begin{equation}
f_{\text {wavelet }}=\operatorname{Concat}(c A, c H, c V, c D)
\end{equation}
These components \( cA \) (approximation), \( cH \) (horizontal detail), \( cV \) (vertical detail), and \( cD \) (diagonal detail) each emphasize unique spatial frequencies of the image. This enhanced map $\mathsf{f}^{'}$ is utilized to modulate the \( f' \) through the $\mathcal{W}$:
\begin{equation}
\mathsf{f}^{'}=f^{\prime}\otimes \mathcal{F} _\uparrow ( \operatorname{Conv}_{3 \times 3}(\sigma (\operatorname{Conv}_{3 \times 3}(f_{\text {wavelet}}))))
\end{equation}
$\mathcal{F} _\uparrow $ represents the up-sampling layer. The final step involves combining the modulated $\mathsf{f}^{'}$ with the $f$ and \( f' \) to form an integrated feature map $f_{\text{combined}}$ :
\begin{equation}
f_{\text{combined}} = \operatorname{Concat}(f, f^{\prime}, \mathsf{f}^{'})
\end{equation}

\subsection{SSMs-
based Semantic Consistency Loss}

\begin{algorithm*}[t]
\caption{SSMs-based Semantic Consistency Loss} 
\SetAlgoInsideSkip{15pt} 
\KwIn{Predicted states $\hat{S}$, Target states $S$, Weight $\lambda$, Directions $K=4$} 
\KwOut{Semantic Loss $\mathcal{L}$} 
\BlankLine

Initialize loss: $\mathcal{L} \gets 0$\; 
Extract dimensions $B \times H \times W \times C$ from $\hat{S}, S$\; 

\vspace{5pt}
\tcc{\textcolor{orange}{Extract directional sequences}}
\vspace{5pt}

\ForEach{direction $k \in \{1, 2, 3, 4\}$}{
    Extract directional sequences $\hat{S}_k$ and $S_k$ by unfolding\;
    Initialize hidden states $h_0 \gets 0$ for all batches\;

    \vspace{5pt}
    \tcc{\textcolor{orange}{Perform state-space model (SSM) operations for all timesteps}}
    \vspace{5pt}
    \For{$t \in \{1, \dots, H \times W\}$}{
        \vspace{3pt}
        $h_t \gets \bar{A} h_{t-1} + \bar{B} \hat{S}_k(t)$\; 
        \vspace{3pt}
        $\hat{y}_t \gets C h_t + D \hat{S}_k(t)$\; 
        \vspace{3pt}
        $h_t^{\text{target}} \gets \bar{A} h_{t-1}^{\text{target}} + \bar{B} S_k(t)$\; 
        \vspace{3pt}
        $y_t^{\text{target}} \gets C h_t^{\text{target}} + D S_k(t)$\ 
    }

    \vspace{5pt}
    \tcc{\textcolor{orange}{Aggregate directional outputs}}
    \vspace{5pt}
     $\hat{y}_k \gets \sum_t \hat{y}_t, \quad y_k^{\text{target}} \gets \sum_t y_t^{\text{target}}$\
}

\vspace{5pt}
\tcc{\textcolor{orange}{Combine outputs from all directions}}
\vspace{5pt}

Aggregate directions: $\hat{Y} \gets \frac{1}{K} \sum_{k=1}^{K} \hat{y}_k, \quad Y^{\text{target}} \gets \frac{1}{K} \sum_{k=1}^{K} y_k^{\text{target}}$\;

\vspace{5pt}
\tcc{\textcolor{orange}{Compute pixel-wise L1 loss}}
\vspace{5pt}

\For{$b \in \{1, \dots, B\}$}{
\vspace{3pt}
    Compute pixel-wise difference: $\mathcal{L}_{b, c, h, w}^{\text{diff}} = \hat{Y}_{b, c, h, w} - Y^{\text{target}}_{b, c, h, w}$\;
    \vspace{3pt}
    Compute squared difference: $\mathcal{L}_{b, c, h, w}^{s} = (\mathcal{L}_{b, c, h, w}^{\text{diff}})^2$\;
    \vspace{5pt}
    $\mathcal{L}_b \gets \frac{1}{C \cdot H \cdot W} \sum_{c=1}^{C} \sum_{h=1}^{H} \sum_{w=1}^{W} \mathcal{L}_{b, c, h, w}^{s}$\;
}

\vspace{5pt}
\tcc{\textcolor{orange}{Aggregate batch losses}}
\vspace{5pt}

Compute batch mean loss: $\mathcal{L} \gets \frac{1}{B} \sum_{b=1}^{B} \mathcal{L}_b$\;
\vspace{3pt}
Scale the loss: $\mathcal{L} \gets \lambda \cdot \mathcal{L}$\;

\Return $\mathcal{L}$\
\label{al.1}
\end{algorithm*}

The block-wise processing mechanism in Mamba-based architectures enables efficient sequence modeling but disrupts global spatial coherence, leading to semantic misalignment across partitioned regions. This fragmentation is particularly problematic in IRSR, where maintaining structural consistency and preserving fine details is crucial. To mitigate this issue, we introduce the SSMs-
based semantic consistency loss, which enforces cross-block feature alignment, ensuring that semantically related regions remain consistent throughout reconstruction (see Algorithm.\ref{al.1}).

Unlike conventional loss functions that focus on pixel-level fidelity, our approach explicitly models semantic dependencies across spatially disjoint regions by leveraging state-space modeling for sequential feature propagation. By constructing directional feature trajectories, our method maintains continuity across non-overlapping blocks, effectively countering the disruptions introduced by independent block-wise processing.

Inspired by 2D-SSM methods\cite{liu2024vmamba}, we reformulate Mamba's sequential modeling to ensure structural consistency across fragmented representations. Given an input feature $X \in \mathbb{R}^{H \times W \times C}$, we process it through two parallel pathways: The primary path expands feature channels to $\lambda C$ using a linear transformation, followed by a depth-wise convolution, a SiLU activation $\sigma$, and a 2D-SSM module for sequential modeling. The secondary pathway applies a similar channel expansion and activation process but omits the depth-wise convolution.

Given the predicted state representations $\hat{S}$ and ground-truth states $S$, we first extract directional sequences across four orientations. Each sequence $\hat{S}_k$ and $S_k$ is transformed using state-space dynamics:

\begin{equation}
\resizebox{0.8\columnwidth}{!}{$
\begin{aligned}
& X_1=\operatorname{LN}(\operatorname{2D-SSM}(\sigma (\operatorname{DWConv}(\operatorname{Linear}(X))))) \\
& X_2=\sigma(\operatorname{Linear}(X)) \\
& X_{\text{out}}=\operatorname{Linear}(X_1 \otimes  X_2)
\end{aligned}
$}
\end{equation} where $h_t$ represents the latent state, and $\hat{y}_t$ is the generated feature output at time step $t$. A similar operation is applied to the ground-truth sequences, yielding target-aligned features $y_t^{\text {target }}$. The feature representations are then aggregated across all directions:

\begin{equation} \hat{Y}=\frac{1}{K} \sum_{k=1}^K \hat{y}k, \quad Y^{\mathrm{target}}=\frac{1}{K}\sum_{k=1}^K y_k^{\mathrm{target}} \end{equation}

To enforce semantic consistency across spatially fragmented regions, we compute the final weighted semantic consistency loss as the L1 distance between the predicted and ground-truth feature representations:

\begin{equation}
\mathcal{L}=\lambda \cdot \frac{1}{B} \sum_{b=1}^B \frac{1}{C \cdot H \cdot W} \sum_{c, h, w}\left(\hat{Y}_{b, c, h, w}-Y_{b, c, h, w}^{\mathrm{target}}\right)^2
\end{equation} where $\lambda$ is a weighting factor that controls the contribution of semantic alignment in the loss function.

By enforcing structural consistency across block-wise feature representations, the proposed loss significantly improves texture preservation and semantic alignment. Unlike conventional losses that rely on pixel-wise differences, this approach explicitly maintains spatial coherence across fragmented regions, mitigating the misalignment introduced by Mamba’s block-wise computation.

\section{Experiment Results \& Details}

\subsection{Datasets and Evaluation Metrics}

\subsubsection{Datasets} To evaluate the effectiveness of IRSRMamba, we utilized a combination of benchmark infrared image datasets that comprehensively assess the model’s performance across diverse imaging conditions. The M3FD dataset\cite{liu2022target}served as the primary HR dataset, from which 265 high-quality infrared images were selected. Their corresponding LR versions were generated via bicubic downsampling, ensuring a controlled degradation process for fair comparisons with existing methods.

For benchmarking, we assessed model performance on two widely used datasets in IRSR research: result-A\cite{liu2018infrared, huang2021infrared}and result-C\cite{zhang2017infrared, huang2021infrared}. These datasets include a diverse range of infrared scenes, covering various target types, imaging conditions, and background complexities, making them well-suited for validating super-resolution models. Additionally, to test the generalization capability of our approach, we employed the CVC10 dataset\cite{campo2012multimodal}, which consists of multi-modal infrared images captured under varying environmental conditions. Evaluating IRSRMamba on CVC10 provides insights into its robustness across different infrared imaging scenarios.

\subsubsection{Evaluation Metrics} 

To provide a rigorous and holistic assessment of the reconstructed image quality, we employed a combination of full-reference image quality metrics and no-reference perceptual quality metrics. The full-reference metrics included Peak Signal-to-Noise Ratio (PSNR), Mean Squared Error (MSE), and Structural Similarity Index (SSIM), which measure the fidelity of the reconstructed images against their ground-truth HR counterparts.

Peak Signal-to-Noise Ratio (PSNR, dB) quantifies the reconstruction fidelity by comparing the pixel-wise intensity differences between the super-resolved and ground-truth images. It is defined as:

\begin{equation}
\operatorname{PSNR}=10 \log _{10}\left(\frac{\max \left(I_{\mathrm{GT}}\right)^2}{\operatorname{MSE}}\right)
\end{equation} where $I_{\mathrm{GT}}$ represents the ground-truth HR image, and MSE denotes the mean squared error, calculated as:

\begin{equation}
\mathrm{MSE}=\frac{1}{N} \sum_{i=1}^N\left(I_{\mathrm{SR}}(i)-I_{\mathrm{GT}}(i)\right)^2
\end{equation} where $I_{\mathrm{SR}}(i)$ is the SR image, $N$ is the total number of pixels, and a lower MSE corresponds to better reconstruction accuracy.

The structural Similarity Index (SSIM) is used to measure perceptual image quality by considering structural, luminance, and contrast similarities between images. It is formulated as:

\begin{equation}
\resizebox{0.8\hsize}{!}{$
\operatorname{SSIM}\left(I_{\mathrm{SR}}, I_{\mathrm{GT}}\right)=\frac{\left(2 \mu_{\mathrm{SR}} \mu_{\mathrm{GT}}+C_1\right)\left(2 \sigma_{\mathrm{SR}, \mathrm{GT}}+C_2\right)}{\left(\mu_{\mathrm{SR}}^2+\mu_{\mathrm{GT}}^2+C_1\right)\left(\sigma_{\mathrm{SR}}^2+\sigma_{\mathrm{GT}}^2+C_2\right)}
$}
\end{equation} where $\mu_{\mathrm{SR}}$ and $\mu_{\mathrm{GT}}$ are the mean intensities of the super-resolved and ground-truth images, respectively, while $\sigma_{\mathrm{SR}}^2$ and $\sigma_{\mathrm{GT}}^2$ denote their variances, and $\sigma_{\mathrm{SR}, \mathrm{GT}}$ represents their covariance. The constants $C_1$ and $C_2$ are stabilization factors to prevent division by zero. Higher SSIM values indicate better perceptual quality and structural fidelity.

Beyond these conventional metrics, we incorporated no-reference perceptual image quality assessment (IQA) methods to evaluate the perceptual realism of the reconstructed images. These methods included Deep Bilinear Convolutional Neural Network (DBCNN)\cite{dbcnn}, Naturalness Image Quality Evaluator (NIQE)\cite{niqe}, Contrastive Language-Image Pretraining for Image Quality Assessment (CLIP-IQA)\cite{clipiqa}, and Blind/Referenceless Image Spatial Quality Evaluator (BRISQUE)\cite{BRISQUE}.

DBCNN is a deep-learning-based approach that predicts perceptual fidelity by leveraging both high-level content features and low-level distortion characteristics, providing a robust assessment of perceptual image quality. NIQE is a model-free statistical measure that quantifies deviations from natural image statistics, where lower scores correspond to higher image quality. CLIP-IQA leverages contrastive learning between vision and language embeddings to provide an assessment aligned with human perception. Finally, BRISQUE analyzes the natural scene statistics of an image to quantify distortions, where lower scores indicate superior image quality.

\subsubsection{Other Settings} All experiments were conducted using PyTorch and executed on an NVIDIA A6000 GPU, ensuring computational efficiency. The training was performed with a batch size of 32 and a learning rate of $1 e^{-5}$, using the Adam optimizer for weight updates.

To maintain evaluation consistency and prevent chrominance artifacts from influencing results, all including IRSRMamba-were evaluated exclusively on the luminance $(\mathrm{Y})$ channel in the YCbCr color space. This standard protocol ensures that color distortions do not affect the accuracy of super-resolution evaluations, making the reported metrics comparable across different methods.

\subsection{Quantitative Results}

\begin{table*}[htbp]
\centering
\renewcommand\arraystretch{1.1}
\caption{The average results of (PSNR$\uparrow$ MSE$\downarrow$ SSIM$\uparrow$) with scale factor of  4 \& 2 on datasets result-A \& result-C \& CVC10. Best and second-best performances are marked in \textbf{bold} and {\ul underlined}, respectively.}
\label{tab:my-table}
\resizebox{0.9\textwidth}{!}{%
\begin{tabular}{@{}c|cccccccccc@{}}
\toprule
\multirow{2}{*}{Scale}       & \multicolumn{1}{c|}{\multirow{2}{*}{Methods}} & \multicolumn{3}{c|}{result-A}                                              & \multicolumn{3}{c|}{result-C}                                              & \multicolumn{3}{c}{CVC10}                             \\ \cmidrule(l){3-11} 
                             & \multicolumn{1}{c|}{}                         & PSNR$\uparrow $  & MSE$\downarrow $ & \multicolumn{1}{c|}{SSIM$\uparrow $} & PSNR$\uparrow $  & MSE$\downarrow $ & \multicolumn{1}{c|}{SSIM$\uparrow $} & PSNR$\uparrow $  & MSE$\downarrow $ & SSIM$\uparrow $ \\ \midrule
\multirow{13}{*}{$\times 2$} & EDSR\textcolor[RGB]{217,205,144}{\textit{[CVPRW 2017]}}~\cite{lim2017enhanced}                                            & 39.0493          & 11.8196          & 0.9414                               & 39.8902          & 8.9865           & 0.9528                               & 44.1770          & 2.7845           & 0.9713          \\
                             & ESRGAN\textcolor[RGB]{217,205,144}{\textit{[ECCVW 2018]}}~\cite{wang2018esrgan}                                        & 38.7738          & 12.5212          & 0.9384                               & 39.6111          & 9.5793           & 0.9500                               & 44.0974          & 2.8477           & 0.9709          \\
                             & FSRCNN\textcolor[RGB]{217,205,144}{\textit{[ECCV 2016]}}~\cite{dong2016accelerating}                                        & 39.1175          & 11.3761          & 0.9426                               & 39.9858          & 8.6899           & 0.9535                               & 44.1253          & 2.8162           & 0.9710          \\
                             & SRGAN\textcolor[RGB]{217,205,144}{\textit{[CVPR 2017]}}~\cite{ledig2017photo}                                        & 39.0401          & 11.9024          & 0.9414                               & 39.8678          & 9.0586           & 0.9527                               & 44.1736          & 2.7851           & 0.9713          \\
                             & SwinIR\textcolor[RGB]{217,205,144}{\textit{[ICCV 2021]}}~\cite{liang2021swinir}                                        & 38.6899          & 12.5694          & 0.9374                               & 39.5215          & 9.6530           & 0.9492                               & 43.9980          & 2.8926           & 0.9704          \\
                             & SRCNN\textcolor[RGB]{217,205,144}{\textit{[T-PAMI 2015]}}~\cite{dong2015image}                                         & 38.9671          & 11.7216          & 0.9414                               & 39.8642          & 8.8857           & 0.9524                               & 44.0038          & 2.9084           & 0.9707          \\
                             & RCAN\textcolor[RGB]{217,205,144}{\textit{[ECCV 2018]}}~\cite{zhang2018image}                                          & 38.8145          & 12.4926          & 0.9391                               & 39.7075          & 9.4220           & 0.9511                               & 44.1205          & 2.8170           & 0.9713          \\
                             & PSRGAN\textcolor[RGB]{217,205,144}{\textit{[SPL 2021]}}~\cite{huang2021infrared}                                       & \underline{39.2146}          & 11.2409          & 0.9429                               & 40.0543          & 8.6101           & 0.9539                               & 44.2377          & 2.7454           & 0.9713          \\
                             & ShuffleMixer(tiny)\textcolor[RGB]{217,205,144}{\textit{[NIPS'22]}}~\cite{sun2022shufflemixer})                            & 39.0465          & 11.7605          & 0.9414                               & 39.8766          & 8.9680           & 0.9527                               & 44.1408          & 2.8113           & 0.9713          \\
                             & ShuffleMixer (base)\textcolor[RGB]{217,205,144}{\textit{[NIPS'22]}}~\cite{sun2022shufflemixer})                            & 38.8066          & 12.3718          & 0.9388                               & 39.6347          & 9.4864           & 0.9503                               & 44.0357          & 2.8809           & 0.9710          \\
                             & HAT\textcolor[RGB]{217,205,144}{\textit{[CVPR 2023]}}~\cite{Chen_2023_CVPR}                                           & 38.7754          & 12.4528          & 0.9384                               & 39.6346          & 9.5132           & 0.9500                               & 44.1080          & 2.8244           & 0.9709          \\
                             & RGT\textcolor[RGB]{217,205,144}{\textit{[ICLR 2024 SOTA]}}~\cite{chen2024recursive}                                              & 39.1642          & 11.3382          & 0.9429                               & 40.0522          & 8.6033           & 0.9540                               & 44.2311          & 2.7358           & 0.9717          \\
                            & \hys{MambaIR}\textcolor[RGB]{217,205,144}{\textit{[ECCV 2024 SOTA]}}~\cite{chen2024recursive}                                              & 39.1761          & \underline{11.2081}          & \underline{0.9437}                               & \underline{40.1399}          & \underline{8.4798}           & \underline{0.9544}                              & \underline{44.4181}          & \underline{2.6076}           & \underline{0.9720}          \\

                             & ATD\textcolor[RGB]{217,205,144}{\textit{[CVPR 2024 SOTA]}}~\cite{zhang2024transcending}                                           & 39.0453          & 11.5702          & 0.9432                               & 40.0375          & 8.6155           & 0.9542                              & 44.1901          & 2.7737           & 0.9711          \\
                             & \textbf{IRSRMamba (Ours)}                                           & \textbf{39.3489} & \textbf{10.8767} & \textbf{0.9440}                      & \textbf{40.2302} & \textbf{8.3164}  & \textbf{0.9548}                      & \textbf{44.5310} & \textbf{2.5537}  & \textbf{0.9720} \\ \midrule
\multirow{13}{*}{$\times 4$} & EDSR\textcolor[RGB]{217,205,144}{\textit{[CVPRW 2017]}}~\cite{lim2017enhanced}                                            & 34.5219          & 30.1273          & 0.8548                               & 35.1740          & 23.9917          & 0.8723                              & 40.1190          & 6.8819           & 0.9482          \\
                             & ESRGAN\textcolor[RGB]{217,205,144}{\textit{[ECCVW 2018]}}~\cite{wang2018esrgan}                                         & 33.6895          & 34.7337          & 0.8500                               & 34.1650          & 28.9017          & 0.8679                               & 37.9780          & 10.9641          & 0.9455          \\
                             & FSRCNN\textcolor[RGB]{217,205,144}{\textit{[ECCV 2016]}}~\cite{dong2016accelerating}                                        & 33.8556          & 34.4909          & 0.8446                              & 34.5272          & 27.4495          & 0.8636                               & 38.7856          & 9.5482           & 0.9421          \\
                             & SRGAN\textcolor[RGB]{217,205,144}{\textit{[CVPR 2017]}}~\cite{ledig2017photo}                                        & 34.5807          & 29.6927         & 0.8556                               & 35.2076          & 23.7701          & 0.8728                               & 40.1479          & 6.8162           & 0.9483          \\
                             & SwinIR\textcolor[RGB]{217,205,144}{\textit{[ICCV 2021]}}~\cite{liang2021swinir}                                        & 34.4321          & 30.6081          & 0.8537                               & 35.0329          & 24.6490          & 0.8710                               & 39.9062          & 7.1886           & 0.9479          \\
                             & SRCNN\textcolor[RGB]{217,205,144}{\textit{[T-PAMI 2015]}}~\cite{dong2015image}                                          & 33.6839          & 34.9181          & 0.8415                               & 34.2348          & 28.6115          & 0.8568                               & 38.0976          & 10.7588          & 0.9279          \\
                             & RCAN\textcolor[RGB]{217,205,144}{\textit{[ECCV 2018]}}~\cite{zhang2018image}                                          & 34.4280          & 30.8815          & 0.8528                               & 35.0823          & 24.6507          & 0.8705                               & 40.0805          & 6.9225           & 0.9484          \\
                             & PSRGAN\textcolor[RGB]{217,205,144}{\textit{[SPL 2021]}}~\cite{huang2021infrared}                                        & 34.4595          & 30.3760          & 0.8540                               & 35.1023          & 24.3147          & 0.8715                               & 39.9533          & 7.1274           & 0.9471          \\
                             & ShuffleMixer(tiny)\textcolor[RGB]{217,205,144}{\textit{[NIPS'22]}}~\cite{sun2022shufflemixer})                            & 34.5440          & 29.9449          & 0.8550                               & 35.1640          & 23.9705          & 0.8723                               & 40.0756          & 6.9296           & 0.9478          \\
                             & ShuffleMixer (base)\textcolor[RGB]{217,205,144}{\textit{[NIPS'22]}}~\cite{sun2022shufflemixer}                            & 34.4507          & 30.6955          & 0.8538                                & 35.0911          & 24.3745          & 0.8714                               & 40.0120          & 7.0622           & 0.9477          \\
                             & HAT\textcolor[RGB]{217,205,144}{\textit{[CVPR 2023]}}~\cite{Chen_2023_CVPR}                                           & 34.4947          & 30.4086          & 0.8542                               & 35.1239          & 24.3103          & 0.8713                               & 40.0934          & 6.9078           & 0.9478          \\
                & RGT\textcolor[RGB]{217,205,144}{\textit{[ICLR 2024 SOTA]}}~\cite{chen2024recursive}                                             & 34.3826          & 31.0046          & 0.8535          & 35.0534          & 24.5924          & 0.8711                & 39.8420          & 7.3060           & 0.9472           \\

                            & MambaIR\textcolor[RGB]{217,205,144}{\textit{[ECCV 2024 SOTA]}}~\cite{chen2024recursive}                                           & 34.0267          & 32.9760          & 0.8510                               & 34.5662          & 27.0850          & 0.8681                               & 38.1878          & 10.8653          & 0.9404          \\
                              & ATD\textcolor[RGB]{217,205,144}{\textit{[CVPR 2024 SOTA]}}~\cite{zhang2024transcending}                                           & \underline{34.6113}          & \underline{29.5294}          & \underline{0.8569}                              & \underline{35.2347}          & \underline{23.6882}          & \underline{0.8737}                              & \underline{40.2897}          & \underline{6.6001}           & \underline{0.9494}      \\
                             & \textbf{IRSRMamba (Ours)}                                     & \textbf{34.6755} & \textbf{29.0551} & \textbf{0.8577}                      & \textbf{35.3074} & \textbf{23.2857} & \textbf{0.8745}                      & \textbf{40.4052} & \textbf{6.4536}  & \textbf{0.9497} \\ \bottomrule
\end{tabular}%
}
\label{tab.2}
\end{table*}

\begin{table*}[t]
\centering
\renewcommand\arraystretch{1.1}
\caption{The average results of (PSNR$\uparrow$ MSE$\downarrow$ SSIM$\uparrow$) with scale factor of  4 \& 2 on datasets result-A \& result-C \& CVC10.}
\label{tab:my-table}
\resizebox{0.9\textwidth}{!}{%
\begin{tabular}{@{}ccccccccccc@{}}
\toprule
\multirow{2}{*}{Scale}                 & \multirow{2}{*}{Methods} & \multicolumn{3}{c}{result-A}                                               & \multicolumn{3}{c}{result-C}                                               & \multicolumn{3}{c}{CVC10}                             \\ \cmidrule(l){3-11} 
                                       &                          & PSNR$\uparrow $  & MSE$\downarrow $ & \multicolumn{1}{c|}{SSIM$\uparrow $} & PSNR$\uparrow $  & MSE$\downarrow $ & \multicolumn{1}{c|}{SSIM$\uparrow $} & PSNR$\uparrow $  & MSE$\downarrow $ & SSIM$\uparrow $ \\ \midrule
\multicolumn{1}{c|}{\multirow{3}{*}{$\times 2$}} & MambaOut\cite{yu2024mambaout}                 & 38.6375          & 12.7091          & 0.9371                               & 39.4900          & 9.7035           & 0.9493                               & 43.9150          & 2.9429           & 0.9704          \\
\multicolumn{1}{c|}{}                  & VisionMamba\cite{zhu2024vision}                   & 38.7805          & 12.2990          & 0.9392                               & 39.6339          & 9.3781           & 0.9506                               & 43.9521          & 2.9103           & 0.9704          \\
\multicolumn{1}{c|}{}              &    \textbf{IRSRMamba (Ours)}                                           & \textbf{39.3489} & \textbf{10.8767} & \textbf{0.9440}                      & \textbf{40.2302} & \textbf{8.3164}  & \textbf{0.9548}                      & \textbf{44.5310} & \textbf{2.5537}  & \textbf{0.9720} \\ \midrule
\multicolumn{1}{c|}{\multirow{3}{*}{$\times 4$}} & MambaOut\cite{yu2024mambaout}                 & 34.4483          & 30.6792          & 0.8527                               & 35.0456          & 24.7055          & 0.8698                               & 40.1587          & 6.8076           & 0.9485          \\
\multicolumn{1}{c|}{}                  & VisionMamba\cite{zhu2024vision}                   & 34.5941          & 29.5650          & 0.8564                               & 35.2327          & 23.6467          & 0.8733                               & 40.1705          & 6.7933           & 0.9484          \\
\multicolumn{1}{c|}{}                 & \textbf{IRSRMamba (Ours)}                                     & \textbf{34.6755} & \textbf{29.0551} & \textbf{0.8577}                      & \textbf{35.3074} & \textbf{23.2857} & \textbf{0.8745}                      & \textbf{40.4052} & \textbf{6.4536}  & \textbf{0.9497}  \\ \bottomrule
\end{tabular}%
}
\label{vmamba}
\end{table*}

To comprehensively assess the performance of IRSRMamba, we conducted quantitative comparisons with both state-of-the-art (SOTA) super-resolution (SR) models and Mamba based backbone architectures. The evaluation was performed on three benchmark datasets-result-A, result-C, and CVC10-at two upscaling factors ( $\times 2$ and $\times 4$ ). The results, summarized in Tables \ref{tab.2} and \ref{vmamba}, utilize PSNR, MSE, and SSIM as evaluation metrics.

\subsubsection{Comparison with State-of-the-Art Super-Resolution Methods} Table \ref{tab.2} presents the performance comparison of IRSRMamba against CNN-, GAN-, and Transformer-based super-resolution models, including EDSR\cite{lim2017enhanced}, ESRGAN\cite{wang2018esrgan}, FSRCNN\cite{dong2016accelerating}, RCAN\cite{zhang2018image}, ATD\cite{zhang2024transcending} and SwinIR\cite{liang2021swinir}. Additionally, recent infrared-specific SR approaches, such as PSRGAN\cite{huang2021infrared} was included for comparison.

For the $\times 2$ upscaling factor, IRSRMamba consistently achieved higher PSNR and SSIM across all datasets, surpassing previous SOTA methods. On the result-A dataset, IRSRMamba achieved a PSNR of 39.35 dB and an SSIM of 0.9440, outperforming ATD (34.04 dB, 0.9432 SSIM) and RGT (39.16 dB, 0.9429 SSIM). On the CVC10 dataset, IRSRMamba reached a PSNR of 44.53 dB, demonstrating its superior ability to reconstruct fine textures in infrared images. The MSE values further confirm the model’s effectiveness, with IRSRMamba achieving an MSE of 10.88 on result-A, significantly lower than MambaIR (11.21) and ATD (11.57).

\subsubsection{Evaluation Against Mamba-Based Architectures} To isolate the impact of IRSRMamba’s architectural innovations, we compared it against other Mamba-based vision models, including MambaOut\cite{yu2024mambaout} and VisionMamba. The results in Table \ref{vmamba} confirm that IRSRMamba outperforms both variants across all datasets.

For the $\times 2$ scale factor, IRSRMamba achieved a PSNR of 39.35 dB on result-A, exceeding VisionMamba (38.78 dB) and MambaOut (38.64 dB). SSIM values followed a similar trend, with IRSRMamba consistently preserving better perceptual quality.

\tgrs{For the $\times 4$ scale factor}, IRSRMamba continued to demonstrate its superiority, particularly on the CVC10 dataset, where it attained a PSNR of 40.41 dB and an SSIM of 0.9497, significantly surpassing VisionMamba (40.17 dB, 0.9484 SSIM) and MambaOut (40.16 dB, 0.9485 SSIM). These results underscore IRSRMamba’s improved feature extraction capabilities and enhanced long-range dependency modeling, enabling it to achieve both high reconstruction fidelity and perceptual realism in infrared image restoration.

\begin{figure*}[hbt!]
    \centering
    \subfigure[Error Distribution Across Models on result-A (Scale $\times 2$)]{
        \begin{minipage}[b]{\textwidth} 
        \centering
        \includegraphics[width=\textwidth]{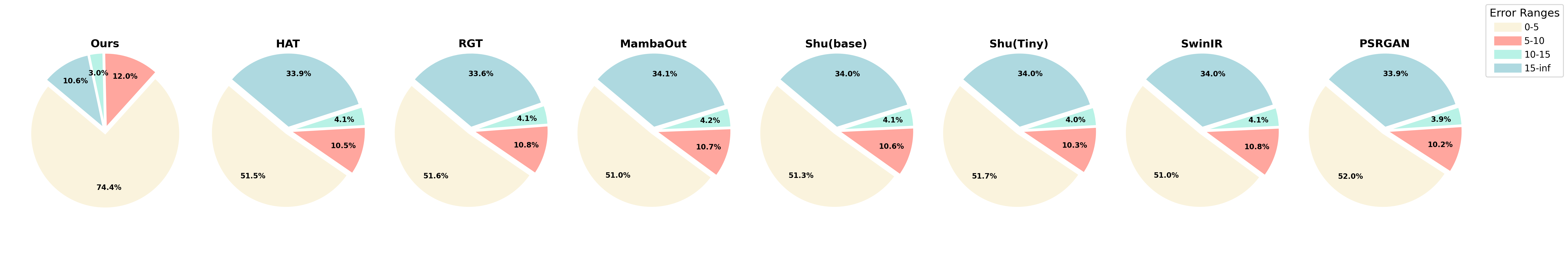} 
        \end{minipage}
        }

    \vspace{1em} 

    \subfigure[Error Distribution Across Models on result-C (Scale $\times 2$)]{
        \begin{minipage}[b]{\textwidth} 
        \centering
        \includegraphics[width=\textwidth]{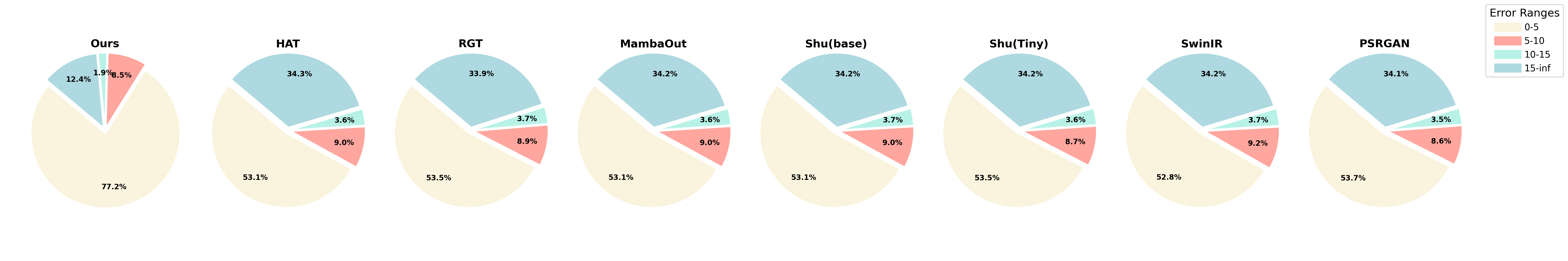} 
        \end{minipage}
        }

    \vspace{1em} 

    \subfigure[Error Distribution Across Models on result-A (Scale $\times 4$)]{
        \begin{minipage}[b]{\textwidth} 
        \centering
        \includegraphics[width=\textwidth]{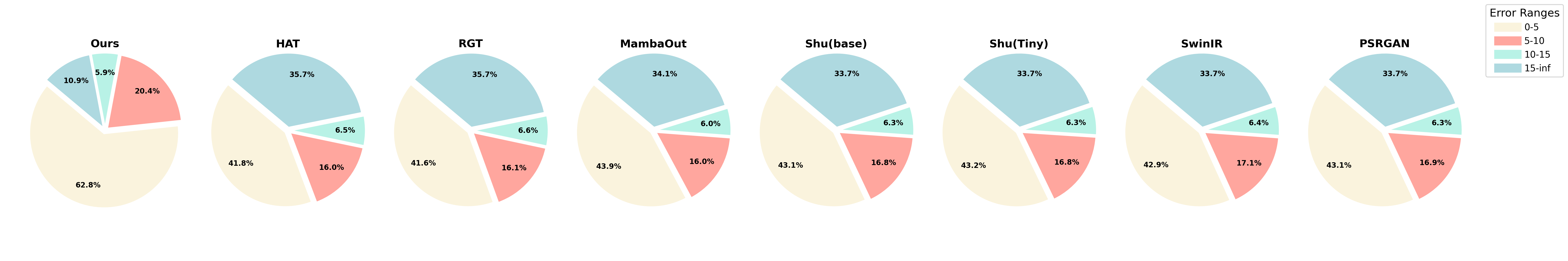} 
        \end{minipage}
        }

    \vspace{1em} 

    \subfigure[Error Distribution Across Models on result-C (Scale $\times 4$)]{
        \begin{minipage}[b]{\textwidth} 
        \centering
        \includegraphics[width=\textwidth]{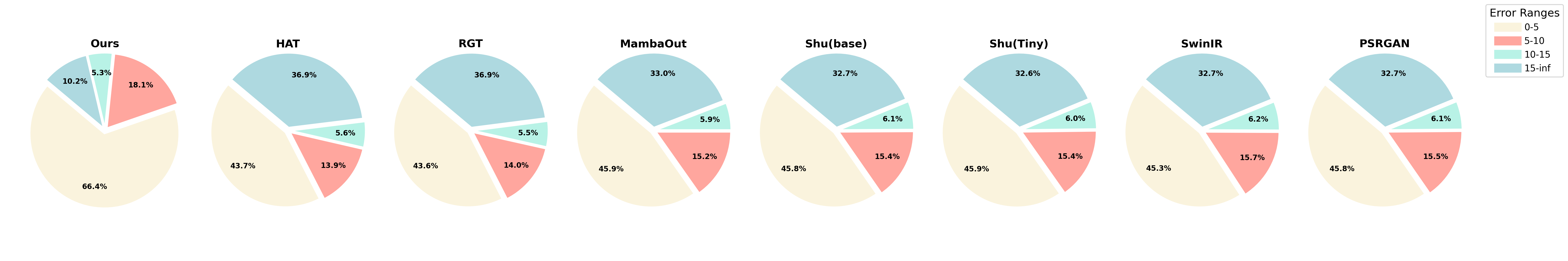} 
        \end{minipage}
        }

    \caption{Global error distribution analysis for various super-resolution models on the result-A and result-C datasets at scaling factors of $\times 2$ and $\times 4$. Each subfigure presents a comparative error breakdown across different ranges, quantifying the proportion of minimal, moderate, high, and severe reconstruction errors.}
    \label{fig:error_distribution_combined}
\end{figure*}

\subsubsection{Global Error Visualization} Quantifying reconstruction errors provides additional insights into how different SR models handle structural details and fine textures, particularly in infrared imaging. Traditional metrics such as PSNR, SSIM, and MSE offer an overall measure of quality but fail to capture localized distortions. To address this limitation, we conduct pixel-wise residual analysis, examining the absolute differences between SR images and their ground truths. Residual errors were categorized into four intensity ranges: \textbf{ranges [0,5) (minimal error), [5,10) (moderate error), [10,15) (high error), and [15, $\infty$ ) (severe error)}. The residual error for each pixel is computed as:

\begin{equation}
E(x, y)=\left|I_{\mathrm{HR}}(x, y)-I_{\mathrm{SR}}(x, y)\right|
\end{equation} where $I_{\mathrm{HR}}(x, y)$ and $I_{\mathrm{SR}}(x, y)$ represent pixel intensity values at coordinate $(x, y)$ in the HR and SR images, respectively. The proportion of pixels falling within each error range is determined using:

\begin{equation}
D_k=\frac{N_k}{N} \times 100 \%
\end{equation} where $D_k$ represents the percentage of pixels in the $k$-th error range, $N_k$ is the number of pixels within that range, and $N$ is the total number of pixels in the image. The overall error distribution is then averaged across all test samples:

\begin{equation}
\bar{D}_k=\frac{1}{M} \sum_{i=1}^M D_k^{(i)}
\end{equation} where $M$ is the total number of test images.

Figure \ref{fig:error_distribution_combined} presents the global error distribution across different models and datasets. Notably, IRSRMamba consistently achieves the highest proportion of minimal-error pixels, highlighting its superior reconstruction accuracy.

For result-A at $\times 2$ scaling, IRSRMamba achieves $74.4 \%$ of pixels in the minimal error range, significantly outperforming PSRGAN (52.0\%) and SwinIR (51.7\%). Similarly, for result-C at $\times 4$ scaling, IRSRMamba maintains $66.4 \%$ minimal error pixels, surpassing HAT (43.7\%) and MambaOut (45.9\%). These findings demonstrate IRSRMamba's ability to suppress artifacts while preserving intricate textures in infrared images.

A closer comparison with Transformer-based models (SwinIR, HAT, and RGT) reveals that, although they achieve competitive PSNR scores, they exhibit higher proportions of high-error pixels. This suggests that while Transformers capture global context, they struggle with localized texture restoration, particularly in high-contrast infrared regions. In contrast, IRSRMamba consistently maintains a lower proportion of high-error pixels, reinforcing its effectiveness in balancing detail preservation and noise suppression.

\subsubsection{Perceptual Quality Assessment} Beyond conventional full-reference metrics (PSNR, SSIM, MSE), perceptual quality assessment is essential for evaluating the visual realism of super-resolved infrared images. Traditional pixel-wise fidelity metrics often fail to align with human perception, particularly in low-contrast or texture-rich regions. To provide a more comprehensive evaluation, we employ no-reference image quality assessment (IQA) methods, including CLIP-IQA(higher is better), DBCNN(higher is better), BRISQUE(lower is better), and NIQE(lower is better). These methods assess both perceptual similarity and distortion artifacts, offering a more holistic perspective on super-resolution performance.  

For a representative analysis, we select the result-A and result-C datasets, which cover a diverse range of infrared imaging conditions and structural complexities. The $\times 2$ upscaling factor is chosen to balance detail preservation and perceptual differences, ensuring a fair comparison across models in realistic application scenarios.

\begin{figure}[htbp]
\centerline{\includegraphics[width=0.8\columnwidth]{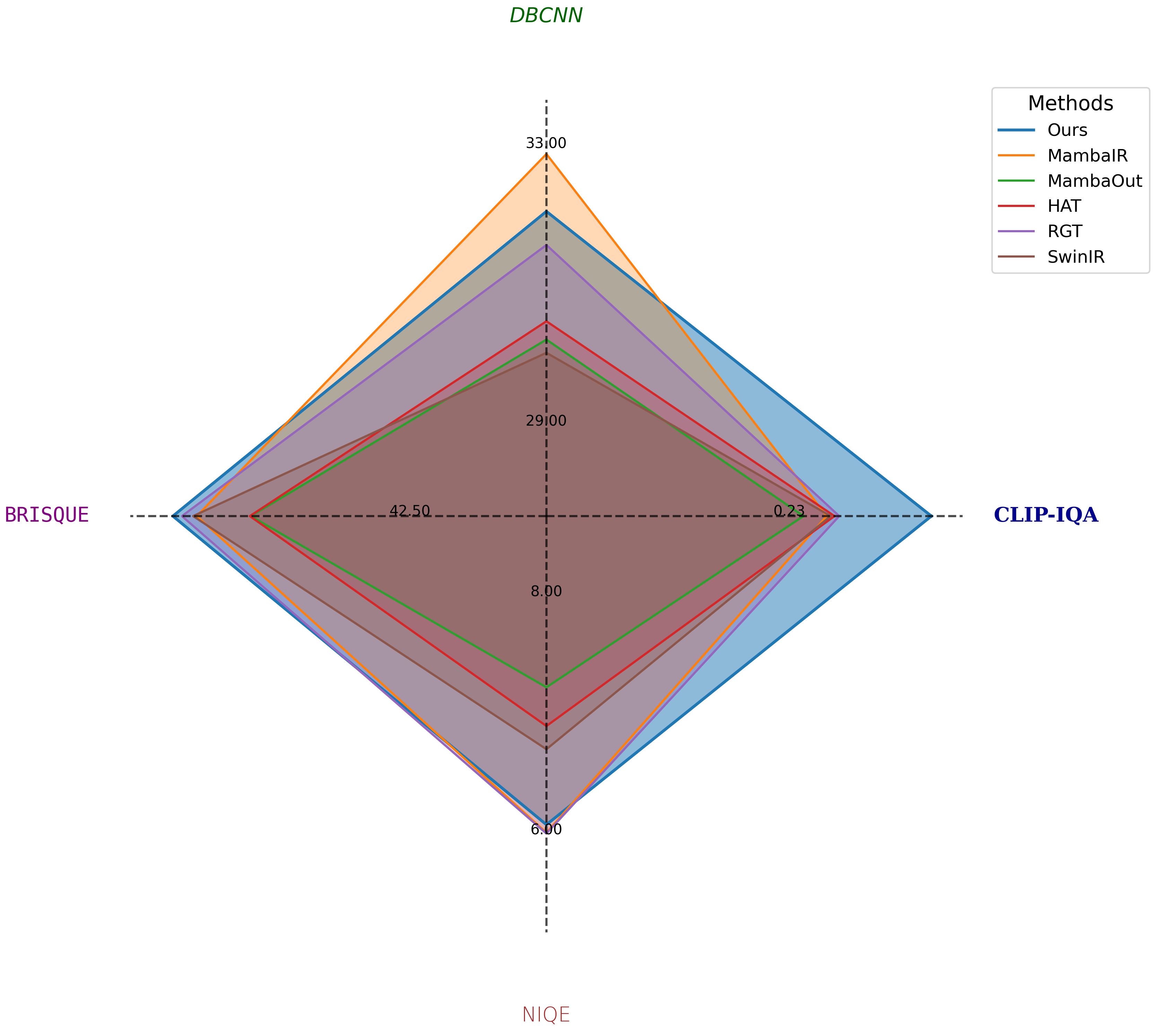}}
\caption{Perceptual quality \tgrs{assessment} of super-resolution models on result-A. Larger area indicates better perceptual performance. (Scale $\times 2$)}
\label{PerceptualA}
\end{figure}

\begin{figure}[htbp]
\centerline{\includegraphics[width=0.8\columnwidth]{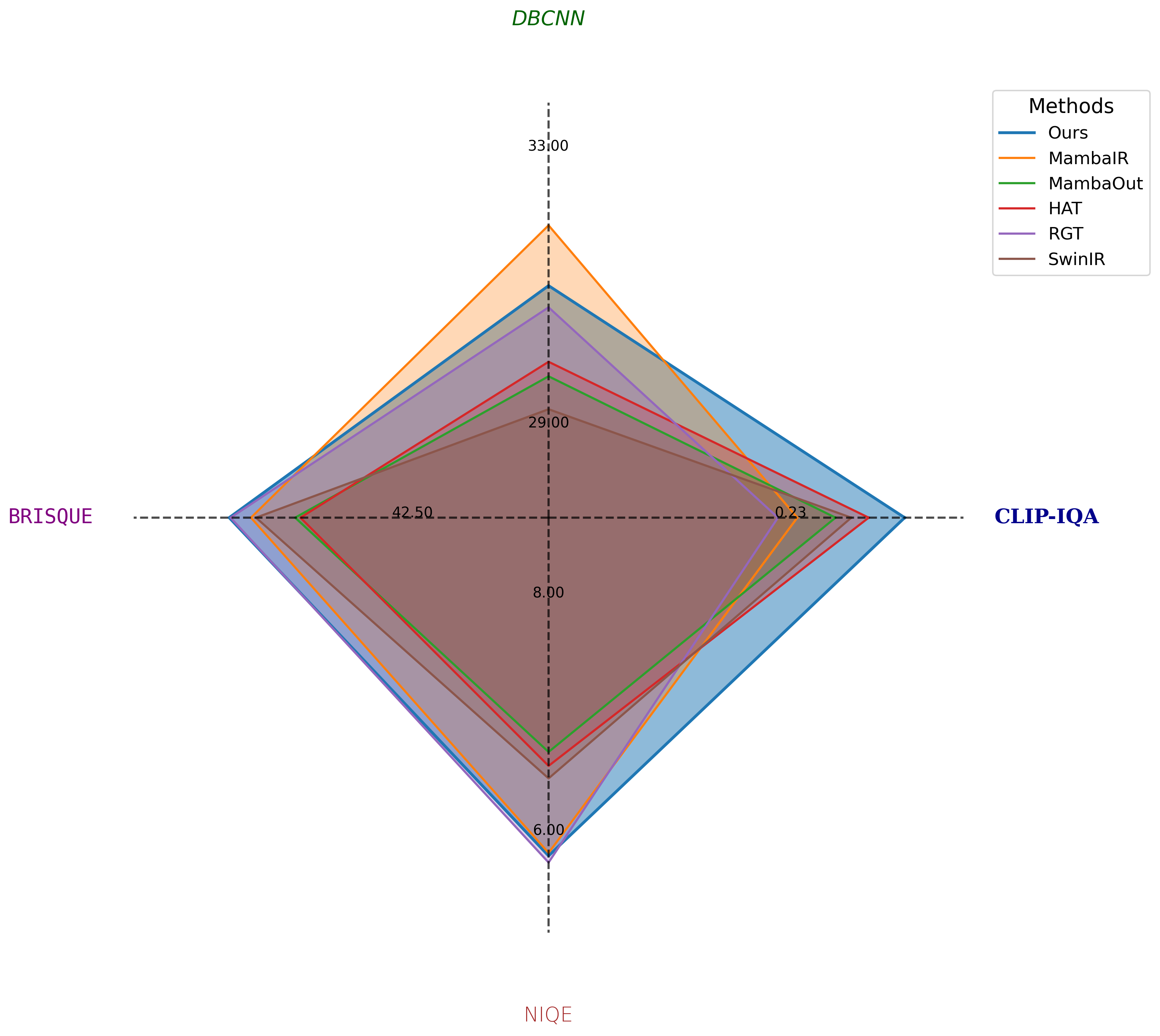}}
\caption{Perceptual quality \tgrs{assessment} of super-resolution models on result-C. Larger area indicates better perceptual performance. (Scale $\times 2$)}
\label{PerceptualC}
\end{figure}

Figures \ref{PerceptualA} and \ref{PerceptualC} present radar chart visualizations of perceptual scores across different super-resolution models for the result-A and result-C datasets, respectively. For result-A, IRSRMamba achieves the highest CLIP-IQA score (0.2887) and DBCNN score (32.3902), reflecting its ability to preserve perceptually significant textures and maintain semantic fidelity. Additionally, it obtains lower BRISQUE (36.9249) and NIQE (5.9063) scores compared to competing models, indicating fewer perceptual artifacts and improved visual quality. In contrast, MambaOut and SwinIR exhibit higher BRISQUE and NIQE values, suggesting that, despite reasonable pixel-wise fidelity, they introduce unnatural distortions that degrade perceptual quality.

A similar trend is observed in result-C, where IRSRMamba consistently outperforms other methods in perceptual quality. The model maintains a higher CLIP-IQA score (0.2787) while achieving lower BRISQUE (38.0739) and NIQE (5.6492) scores, further validating its ability to preserve fine details while minimizing distortions. Comparisons with HAT and RGT reveal that while Transformer-based models achieve competitive DBCNN scores, they exhibit higher NIQE values, suggesting that their outputs may introduce unnatural textures despite capturing global dependencies.


\begin{figure*}[htbp]
\centerline{\includegraphics[width=\textwidth]{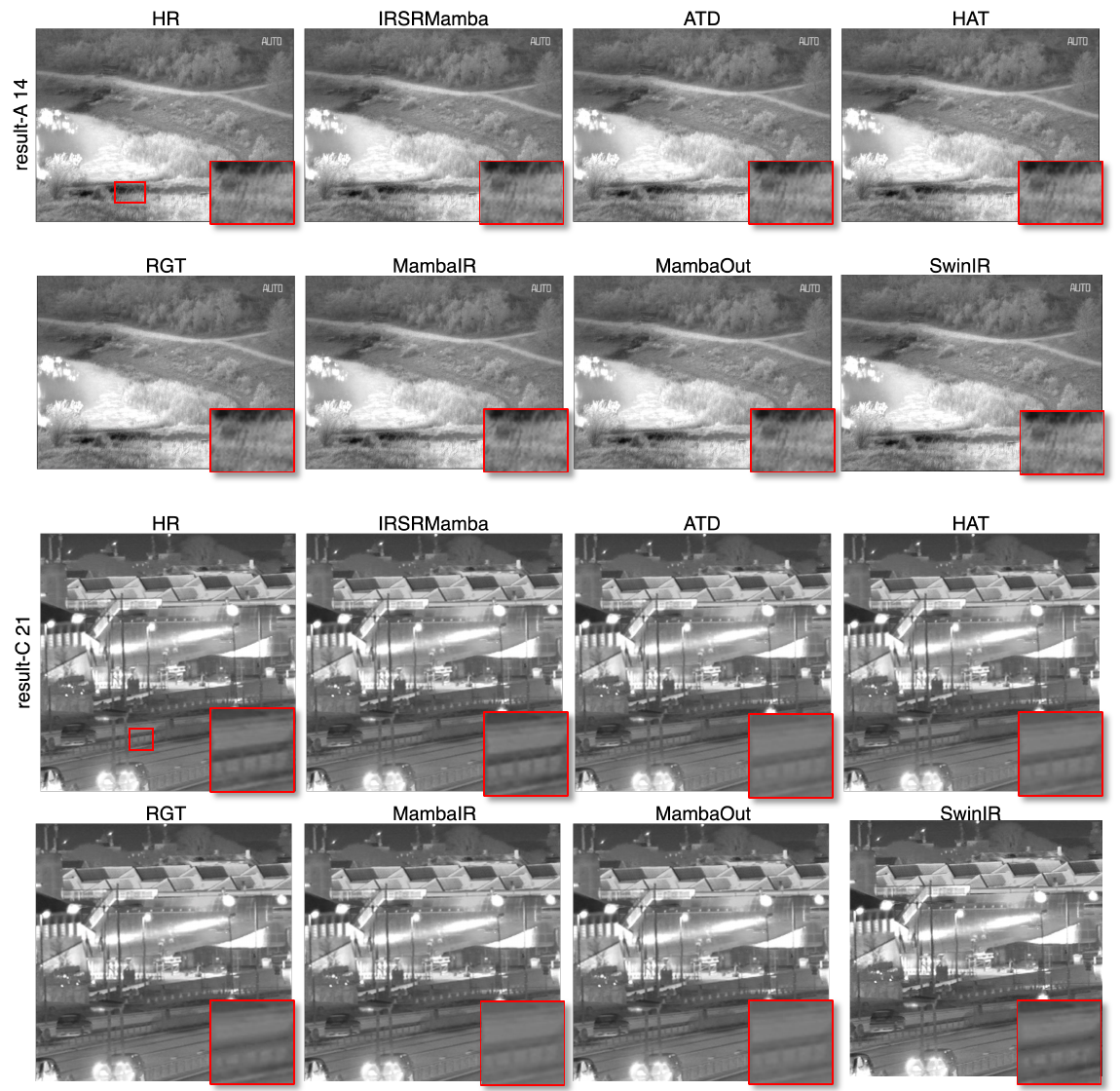}}
\caption{Visual comparison of IRSRMamba and competing models on infrared datasets (result-A and result-C) with a scaling factor of $\times 2$. Each row represents a different dataset, with magnified insets highlighting critical texture regions.}
\label{v_x2}
\end{figure*}

\begin{figure*}[t]
\centerline{\includegraphics[width=\textwidth]{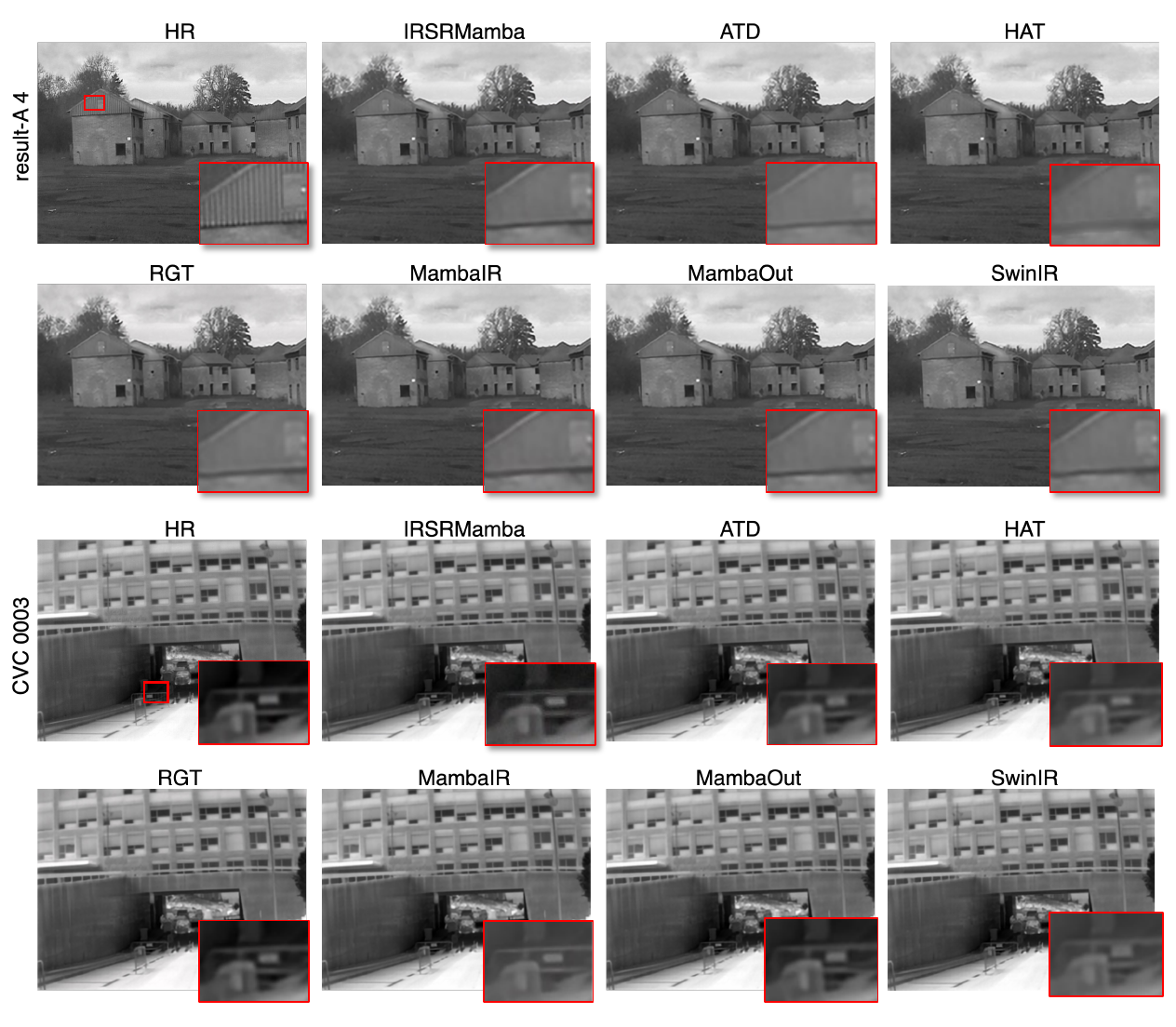}}
\caption{Visual comparison of IRSRMamba and competing models on infrared datasets (result-A and CVC10) with a scaling factor of $\times 4$. Each row represents a different dataset, with magnified insets highlighting critical texture regions.}
\label{reply_x4}
\end{figure*}

\subsection{Qualitative Results} To complement the quantitative assessment, we conduct a subjective visual evaluation by comparing IRSRMamba against state-of-the-art super-resolution models, including ATD, HAT, RGT, SwinIR, MambaIR, and MambaOut. This qualitative analysis provides deeper insights into how well each method preserves structural details, suppresses artifacts, and enhances perceptual quality in infrared images. 

Fig. \ref{v_x2} and \ref{reply_x4} present qualitative comparisons across the result-A, result-C, and CVC10 datasets, evaluated at scaling factors of ×2 and ×4, respectively. Each set consists of the HR reference image, SR image from competing models, and magnified insets that highlight critical texture regions for fine-detail analysis. These enlarged sections allow for a clearer examination of how different models handle fine structures, edges, and noise suppression in infrared imaging. Across all datasets, IRSRMamba consistently demonstrates superior reconstruction quality, effectively retaining sharp edges, intricate textures, and structural integrity compared to other methods. In result-A (Fig. \ref{v_x2}), IRSRMamba reconstructs high-frequency details, such as grass textures and terrain structures, with greater precision, while competing models introduce noticeable blurring and texture loss. SwinIR and MambaOut, for instance, exhibit excessive smoothing, leading to a loss of fine structures, whereas ATD and HAT struggle to capture subtle intensity variations in low-contrast regions, reducing the overall perceptual quality.

A similar pattern emerges in result-C, where IRSRMamba excels in preserving architectural features and small-scale structures. As illustrated in Fig. \ref{v_x2}, the model maintains sharp, well-defined contours in urban environments, while alternative approaches suffer from edge distortions and artifact formation. At a scaling factor of ×4 (Fig. \ref{reply_x4}), IRSRMamba further distinguishes itself by maintaining geometric consistency and reducing the blurry reconstructions commonly observed in transformer-based models such as SwinIR and HAT.

The perceptual evaluation results demonstrate that IRSRMamba consistently outperforms competing models across multiple quality measures, achieving: Higher perceptual similarity (CLIP-IQA, DBCNN), indicating better texture retention and semantic consistency.
Lower distortion artifacts (BRISQUE, NIQE), signifying fewer unnatural artifacts and improved visual realism.
These findings highlight IRSRMamba’s superior ability to balance pixel-wise fidelity and perceptual realism, making it more suitable for real-world infrared imaging applications.

\subsection{Ablation Experiments}

To systematically evaluate the contributions of our proposed enhancements, we conduct ablation experiments by incrementally integrating wavelet modulation and SSM loss into the baseline Mamba architecture. The quantitative results in Tables \ref{tab.1} and \ref{tab.2} confirm that each component significantly improves reconstruction quality across multiple datasets and scaling factors.

\begin{table}[t]
\centering
\renewcommand\arraystretch{1.2}
\caption{Ablation experiment with the performance of different backbone networks across different test datasets ($\times 4$).}
\label{table1}
\resizebox{\columnwidth}{!}{%
\begin{tabular}{@{}c|ccc@{}}
\toprule
    Scale $\times 4$     & Mamba (Pure) & $\to $ \textit{w}/Wavelet Modulation & $\to $ \textit{w}/SSM Loss \\ \midrule
result-A & 34.02/0.8510   & 34.65/0.8569       & 34.67/0.8577          \\
result-C & 34.56/0.8681   & 35.28/0.8738       & 35.30/0.8745          \\
CVC10    & 38.18/0.9404   & 40.38/0.9492       & 40.40/0.9497         \\ \bottomrule
\end{tabular}%
}
\label{tab.2}
\vspace{-8pt}
\end{table}

\begin{table}[t]
\centering
\renewcommand\arraystretch{1.2}
\caption{Ablation experiment with the performance of different backbone networks across different test datasets ($\times 2$).}
\label{table1}
\resizebox{\columnwidth}{!}{%
\begin{tabular}{@{}c|ccc@{}}
\toprule
    Scale $\times 2$     & Mamba (Pure) & $\to $ \textit{w}/Wavelet Modulation & $\to $ \textit{w}/SSM Loss \\ \midrule
result-A & 38.66/0.9372   & 39.33/0.9439       & 39.34/0.9440        \\
result-C & 39.49/0.9487   & 40.21/0.9548       & 40.23/0.9548          \\
CVC10    & 43.98/0.9704   & 44.50/0.9719       & 44.53/0.9720     \\ \bottomrule
\end{tabular}%
}
\label{tab.1}
\vspace{-8pt}
\end{table}

\subsubsection{Effect of Wavelet Modulation and SSM Loss}

Wavelet modulation enhances multi-scale feature extraction, leading to notable improvements in PSNR and SSIM, while the SSM loss further refines structural coherence by enforcing perceptual consistency. For instance, on the CVC10 dataset at $\times 4$ scaling, the baseline Mamba (Pure) achieves 38.18 dB PSNR / 0.9404 SSIM, which improves to 40.38 dB PSNR / 0.9492 SSIM with wavelet modulation, and further increases to 40.40 dB PSNR / 0.9497 SSIM when SSM loss is incorporated. Similar trends are observed across Result-A and Result-C, demonstrating that both components play a crucial role in preserving fine textures and structural details.

\begin{figure}[t]
\centerline{\includegraphics[width=0.9\columnwidth]{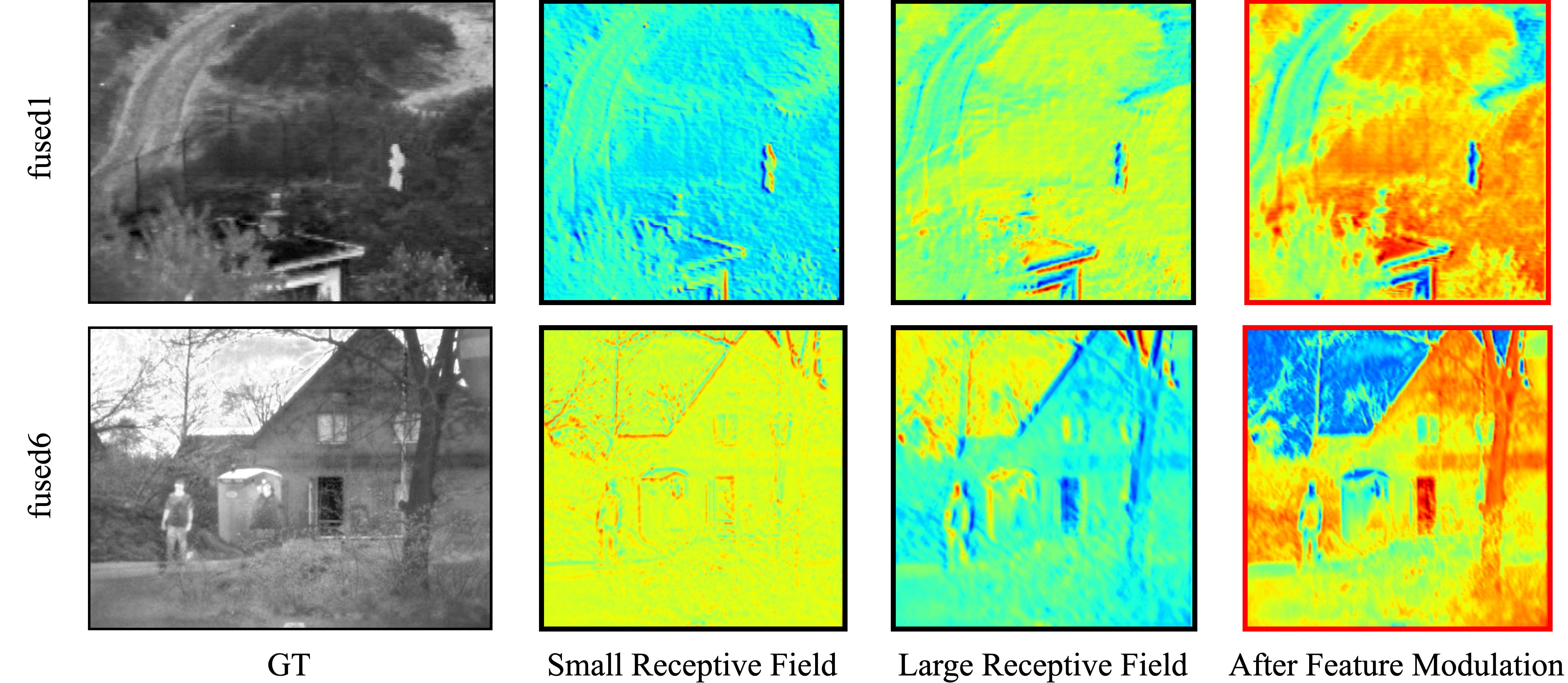}}
\caption{Feature map visualization comparison. From left to right: GT from result-A dataset, feature maps applied with the $3 \times 3$ convolutional kernel, feature maps from the $7 \times 7$ convolutional kernel, feature maps after wavelet transform feature modulation.}
\label{fig.4}
\vspace{-12pt}
\end{figure}

\begin{figure}[htbp]
\centerline{\includegraphics[width=\columnwidth]{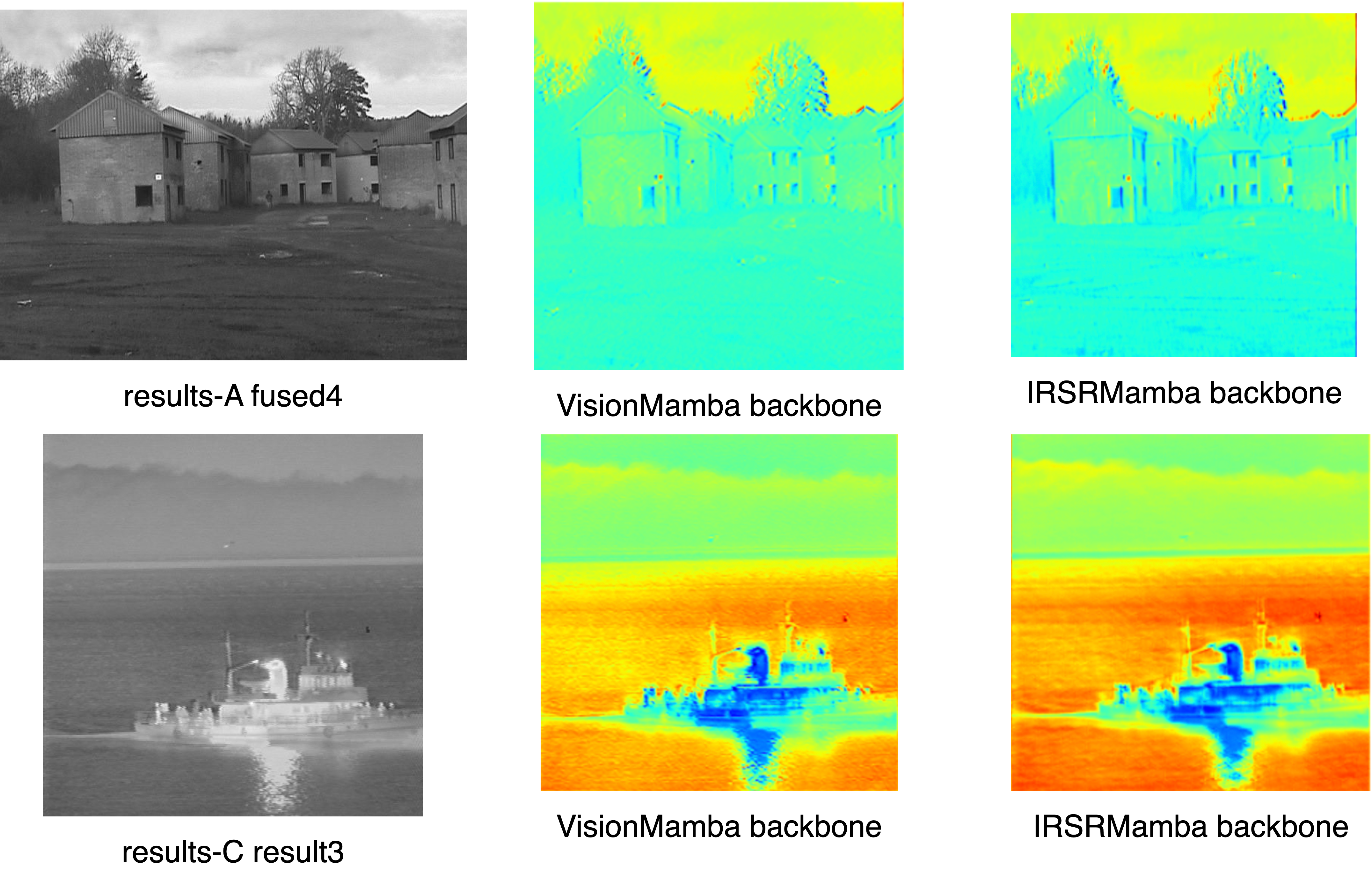}}
\caption{Comparative visualization of feature maps for infrared super-resolution using VisionMamba and IRSRMamba backbones.}
\label{visualization}
\end{figure}

To visually illustrate these enhancements, Figure \ref{fig.4} presents activation visualizations before and after module integration, highlighting a clear improvement in the network's ability to capture globally significant information. The results confirm that wavelet modulation effectively enhances reconstruction fidelity and perceptual quality, making IRSRMamba a robust solution for infrared image super-resolution.

Further validation is provided by feature activation comparisons between IRSRMamba and VisionMamba. As shown in Figure \ref{visualization}, IRSRMamba exhibits stronger structural awareness and higher feature contrast, enabling superior reconstruction of fine details compared to VisionMamba.

\begin{table}[t]
\renewcommand\arraystretch{1.2}
\caption{Ablation analysis of differ with different components. the best PSNR and MSE performance is shown in bold.}
\resizebox{\columnwidth}{!}{%
\begin{tabular}{@{}c|c|ccc@{}}
\toprule
Block Numbers & Param.(K) & \multicolumn{1}{c|}{result-A} & \multicolumn{1}{c|}{result-C} & CVC10          \\ \midrule
4     & 14,382    & 39.3160/10.9495               & 40.2292/8.3347                & 44.5508/2.5492 \\
6     & 20,421    & 39.3164/10.9534               & 40.2383/8.3169                & 44.5534/2.5461 \\
8     & 26,462    & 39.3200/10.9498               & 40.2408/8.3084                & 44.5503/2.5462 \\
10    & 32,502    & 39.3194/10.9506               & 40.2404/8.3094                & 44.5500/2.5465 \\ \bottomrule
\end{tabular}%
}
\label{tab:ablation_blocks}
\end{table}

\begin{figure}[t]
\centerline{\includegraphics[width=\columnwidth]{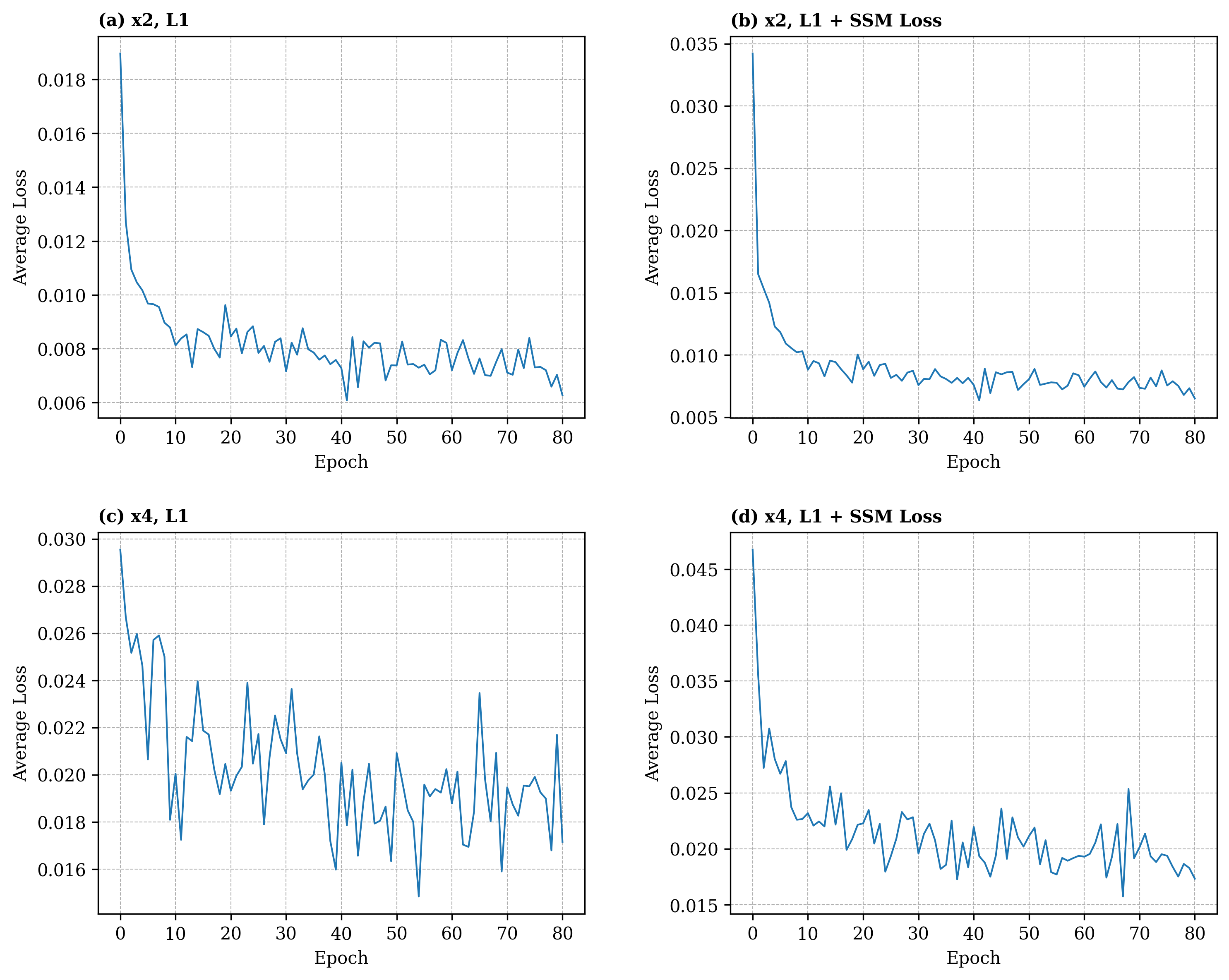}}
\caption{Comparative analysis of average training loss trends by epoch across different sampling scales and loss functions.}
\label{reply_x2}
\end{figure}

\begin{figure}[t]
    \centering
    \subfigure[PSNR and SSIM comparison on the result-A dataset ($\times 2$).]{
        \begin{minipage}[b]{\columnwidth} 
        \centering
        \includegraphics[width=\columnwidth]{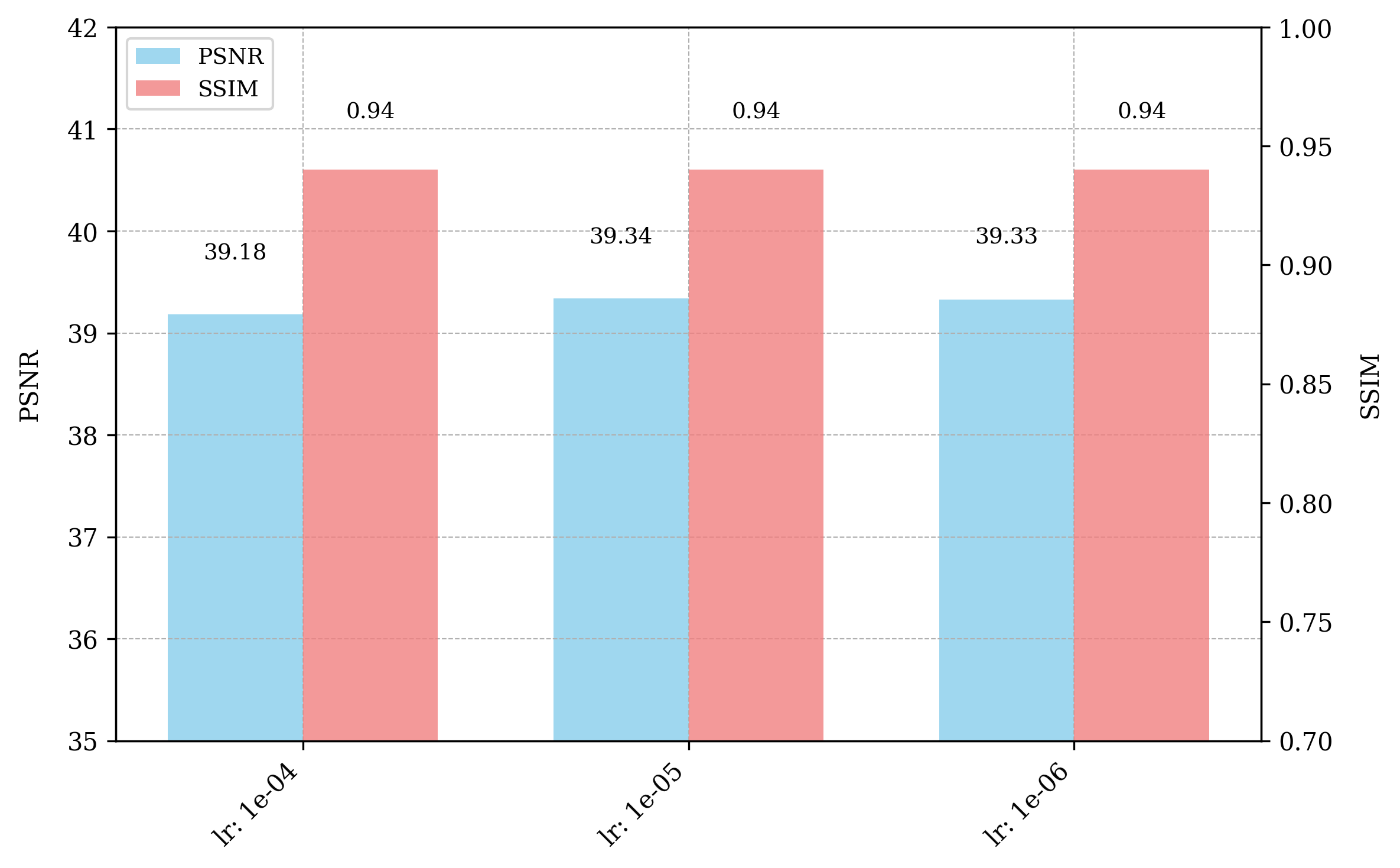} 
        \end{minipage}
        }

    \vspace{0.5em} 

    \subfigure[PSNR and SSIM comparison on the result-C dataset ($\times 2$).]{
        \begin{minipage}[b]{\columnwidth} 
        \centering
        \includegraphics[width=\columnwidth]{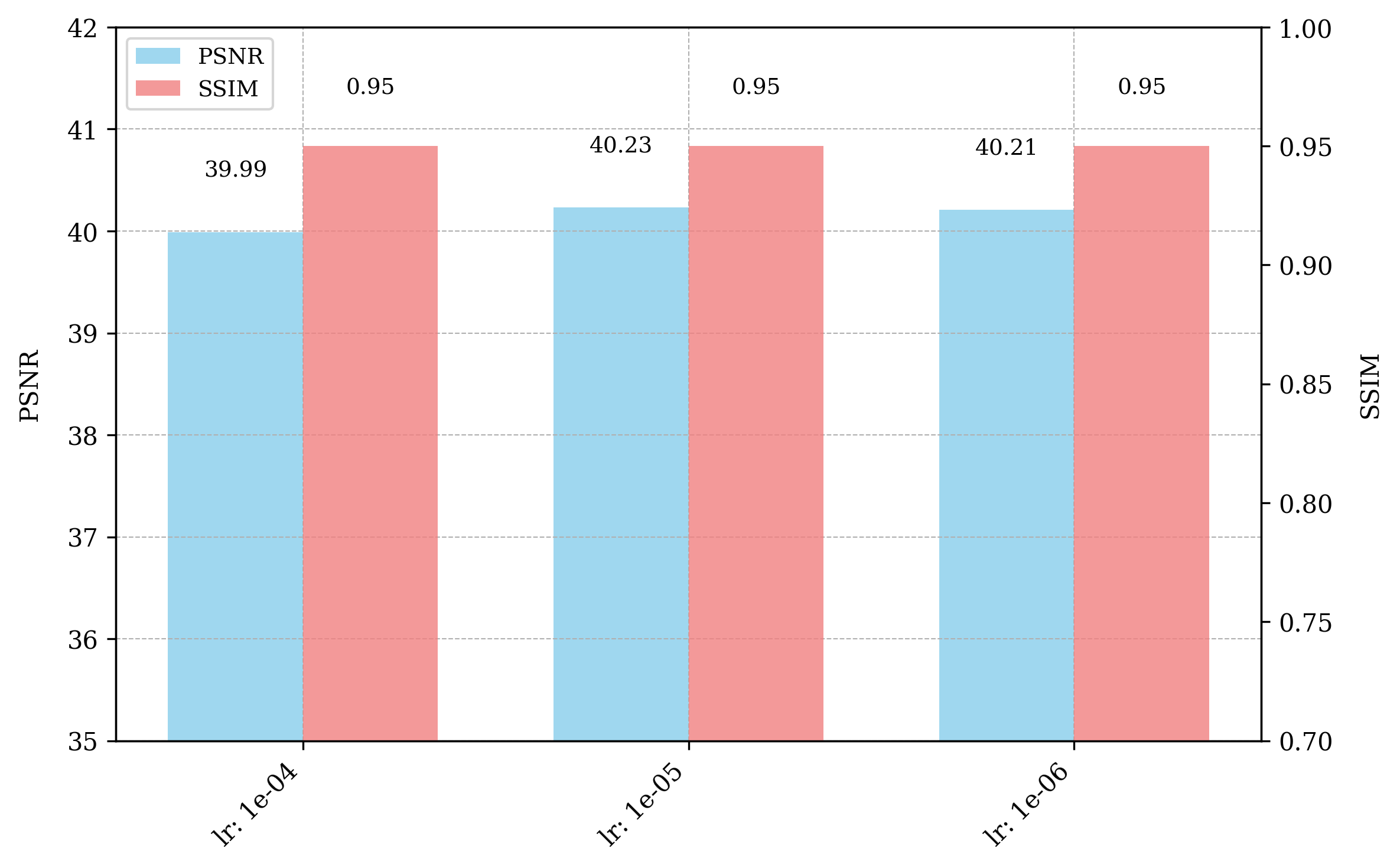} 
        \end{minipage}
        }

    \caption{PSNR and SSIM comparison across result-A and result-C datasets (upsampling factor: $\times 2$). Each subfigure displays the best performance achieved during the first 400 training epochs under different learning rates.}
    \label{fig:overall}
\end{figure}

\subsubsection{Effect of Module Quantity} To determine the optimal number of blocks in IRSRMamba, we evaluate its performance with different block configurations. As shown in Table \ref{tab:ablation_blocks}, increasing the number of blocks from 4 to 8 leads to a progressive improvement in PSNR and MSE. However, further increasing the number to 10 blocks provides negligible additional gains. For example, on the result-C dataset, PSNR increases from 40.2292 dB (4 blocks) to 40.2408 dB (8 blocks), but no further benefit is observed at 10 blocks.

These findings indicate that 8 blocks provide the best balance between reconstruction quality and computational efficiency, maximizing performance without introducing unnecessary computational overhead.

\subsubsection{Effect of Loss Function Composition} We analyze the impact of different loss function compositions on training stability and reconstruction performance. As shown in Fig. \ref{reply_x2}, models trained with L1 loss alone exhibit greater fluctuations, while integrating SSM loss leads to smoother convergence and lower final loss values.

For $\times 2$ super-resolution (Fig. \ref{reply_x2}a-b), the addition of SSM loss accelerates convergence and enhances structural consistency. At $\times 4$ scaling (Fig. \ref{reply_x2}c-d), SSM loss further stabilizes training, reducing variability in high-magnification scenarios. These results confirm that SSM loss enhances feature regularization, improving overall reconstruction fidelity and making it a key component of IRSRMamba.

\subsubsection{Impact of Learning Rate} We investigate the influence of different learning rates on PSNR and SSIM to determine the optimal training setting. As shown in Figure \ref{fig:overall}, learning rate adjustments significantly affect reconstruction performance across datasets.

For result-A (Fig. \ref{fig:overall}a), increasing the learning rate from $1 e^{-6}$ to $1 e^{-5}$ improves PSNR and SSIM, but further increasing it to $1 e^{-4}$ results in diminishing returns. A similar trend is observed in result-C (Fig. \ref{fig:overall}b), where a learning rate of $1 e^{-5}$ provides the best trade-off between accuracy and stability.


\section{Conclusion}

In this study, we propose IRSRMamba, a novel infrared image super-resolution model that integrates Mamba-based state-space modeling, wavelet-driven feature modulation, and SSMs-based semantic consistency loss to enhance long-range dependency modeling, multi-scale feature extraction, and structural coherence. By addressing the limitations of block-wise processing and enforcing cross-block feature alignment, IRSRMamba achieves state-of-the-art performance across benchmark datasets, surpassing existing methods in PSNR, SSIM, and MSE. The incorporation of wavelet modulation improves multi-scale representation, while SSM loss enhances perceptual consistency, ensuring superior texture preservation and artifact suppression. Our findings establish Mamba architectures as a powerful alternative for infrared image enhancement, offering new directions for future research in state-space modeling for super-resolution, denoising, and infrared scene understanding.

\section*{Acknowledgments}
This work was supported by JSPS KAKENHI Grant Number JP23KJ0118.

\ifCLASSOPTIONcaptionsoff
  \newpage
\fi

\begin{IEEEbiography}[{\includegraphics[width=1in,height=1.25in,clip,keepaspectratio]{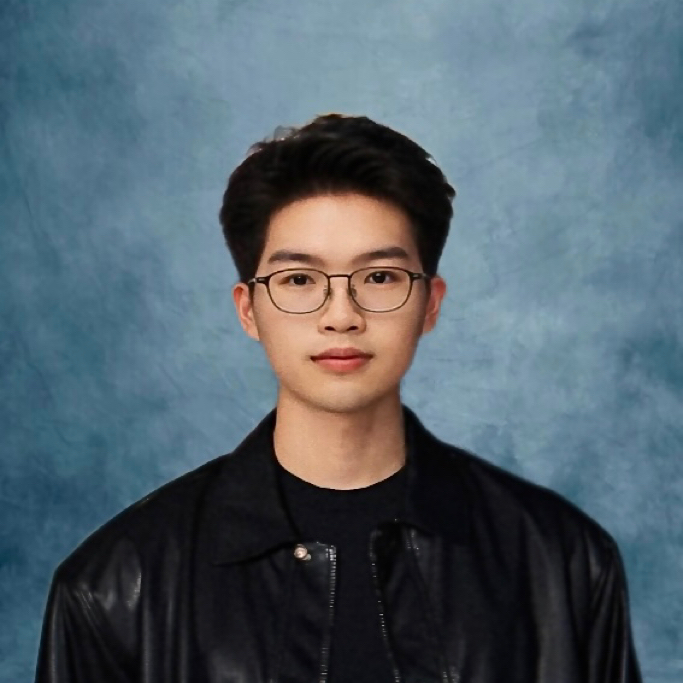}}]{Yongsong Huang}(Member, IEEE)
is an Assistant Professor at the Center for So-Go-Chi (Convergence Knowledge）, Tohoku University. He received his B.E. and M.E. degrees from Guilin University of Electronic Technology in 2018 and 2021, respectively, and his Ph.D. degree from Tohoku University, Japan, in 2025. He has served as a JSPS Research Fellow and has held visiting researcher positions at Harvard University, Massachusetts General Hospital, and National Taiwan University. He is a co-author of the book ‘Applications of Generative AI’ published by Springer. In 2024, he was awarded the JSPS Special Research Grant Allowance. His research interests include computer vision, medical image processing, and image super-resolution.
\end{IEEEbiography}

\begin{IEEEbiography}[{\includegraphics[width=1in,height=1.25in,clip,keepaspectratio]{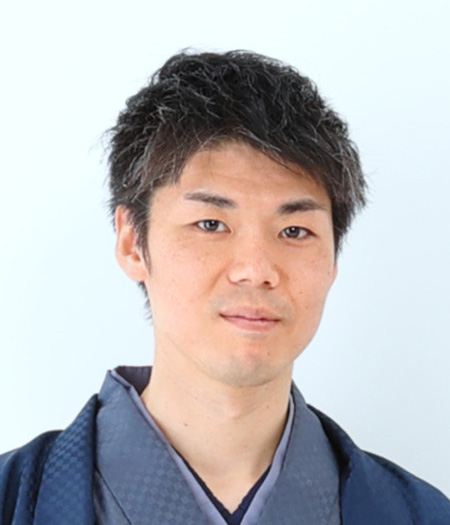}}]{Tomo Miyazaki} 
(Member, IEEE) received B.E. and Ph.D. degrees from Yamagata University and Tohoku University in 2006 and 2011, respectively. He worked on a Geographic Information System at Hitachi, Ltd until 2013. He was a postdoctoral researcher and an assistant professor from 2013 to 2023 at Tohoku University. Since 2024, he has been an Associate Professor. His research interests include pattern recognition and image processing.
\end{IEEEbiography}


\begin{IEEEbiography}[{\includegraphics[width=1in,height=1.25in,clip,keepaspectratio]{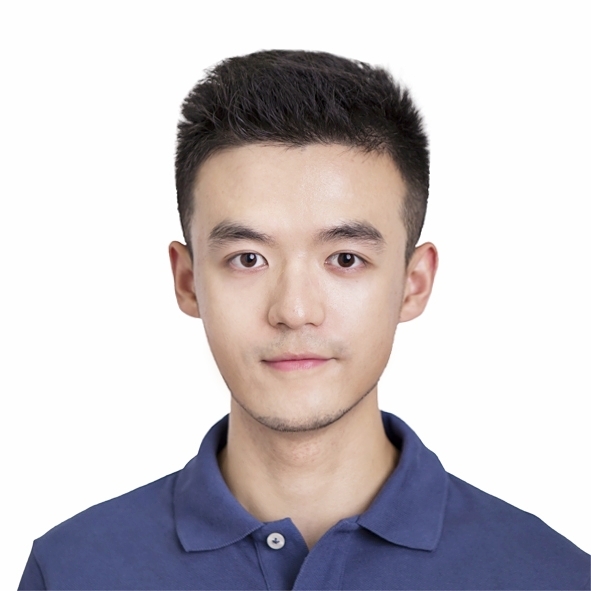}}]{Xiaofeng Liu} (Member, IEEE) is an Assistant Professor at Yale, and an Associate Member at the Broad Institute of MIT and Harvard. He is also an affiliate faculty member of the Center for Biomedical Data Science, Yale Institutes for Foundations of Data Science, and Global Health. His Ph.D. was jointly supervised by advisors from University of Chinese Academy of Science and Carnegie Mellon University. He received the B.Eng. and B.A. from the University of Science and Technology of China in 2014. He was a recipient of the Best Paper award of the IEEE ISBA 2018. He is the program committee or reviewer for PAMI, TIP, TNNLS, NeurIPS, ICML, CVPR, ICCV, ECCV, Nat. Com. His research interests are centered around the convergence of trustworthy AI/deep learning, medical imaging, and data science to advance the diagnosis, prognosis, and treatment monitoring of various diseases.
\end{IEEEbiography}

\begin{IEEEbiography}[{\includegraphics[width=1in,height=1.25in,clip,keepaspectratio]{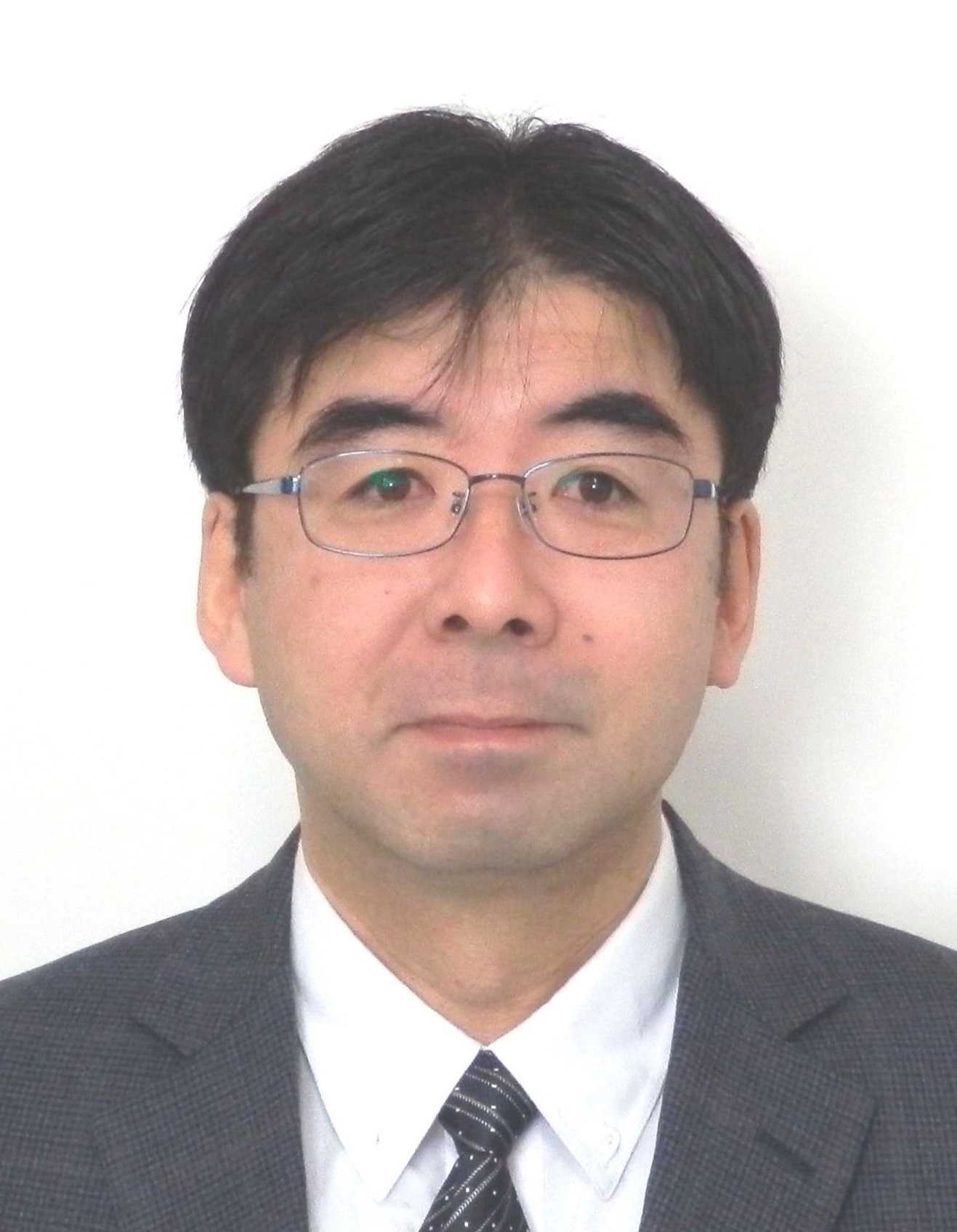}}]{Shinichiro Omachi} 
(Senior Member, IEEE) received the B.E., M.E., and Ph.D. degrees in information engineering from Tohoku University, Japan, in 1988, 1990, and 1993, respectively. He worked as an Assistant Professor at the Education Center for Information Processing,
Tohoku University, from 1993 to 1996. Since 1996, he has been affiliated with the Graduate School of Engineering, Tohoku University, where he is currently a Professor. From 2000 to 2001, he was a Visiting Associate Professor at Brown University.
His research interests include pattern recognition, computer vision, image processing, image coding, and parallel processing. He is a member of the Institute of Electronics, Information and Communication Engineers, the Information Processing Society of Japan, among others. He received the IAPR/ICDAR Best Paper Award in 2007, the Best Paper Method Award of the 33rd Annual Conference of the GfKl in 2010, the ICFHR Best Paper Award in 2010, and the IEICE Best Paper Award in 2012. From 2020 to 2021, he was the Vice Chair of the IEEE Sendai Section. He served as the Editor-in-Chief for IEICE Transactions on Information and Systems from 2013 to 2015.
\end{IEEEbiography}

\balance
\section*{}
\bibliographystyle{ieeetr} 
\bibliography{IEEEabrv,ref}

\end{document}